\newif\ifnotes
\notestrue 

\documentclass[10pt, fleqn]{article}

\usepackage[usenames,dvipsnames,table]{xcolor}
\usepackage{multicol} 

\usepackage{wrapfig}
\usepackage{algpseudocode}

\usepackage[normalem]{ulem}
\usepackage{setspace}
\usepackage{siunitx} 
\usepackage{pgf}
\usepackage{tikz}
\usepackage{tikzscale}
\usepackage{IEEEtrantools}
\usetikzlibrary{arrows,automata, calc, decorations.pathreplacing, shapes,positioning}
\usepackage{subcaption} 
\ifnotes
  \usepackage[nomargin,inline,draft]{fixme}
\else
  \usepackage[nomargin,inline,final]{fixme}
\fi
\usepackage{todonotes}
\usepackage{makecell}

\usepackage[utf8]{inputenc}
\usepackage[T1]{fontenc}
\usepackage{microtype}
\usepackage{pstricks}
\usepackage{graphicx}
\usepackage{amsmath}
\usepackage{amssymb}
\usepackage{xspace}
\usepackage{listings}
\usepackage{multirow} 
\usepackage{amsthm}

\usepackage[top=1.5cm, bottom=2cm, left=2cm, right=2cm]{geometry}

\usepackage{placeins}

\usepackage{mathtools}

\usepackage{float}

\usepackage{flushend}

\usepackage[
  sorting=none,
  style=numeric-comp,
  backend=biber,
  isbn=false,
  url=true,
  doi=false,
  url=false,
  mincrossrefs=100,
  maxnames=12,
  giveninits=true
]{biblatex}

\AtEveryBibitem{\clearname{editor}}

\definecolor{vrpink}{RGB}{255,0,127}
\definecolor{vrblue}{RGB}{30,144,255}
\definecolor{vrolive}{RGB}{85,107,47}
\definecolor{vrroyalblue}{RGB}{65,105,225}
\definecolor{brgreen}{RGB}{100,200,70}
\definecolor{ivsalmon}{RGB}{255,160,122}
\definecolor{vrlpink}{RGB}{255,192,203}
\definecolor{mvcol}{RGB}{5,150,25}

\definecolor{jdgreen}{RGB}{4,99,7}
\definecolor{jdred}{RGB}{133,6,6}

\fxusetheme{color}
\fxuseenvlayout{color}

\usepackage{graphicx}
\graphicspath{{./figures/}}
\DeclareGraphicsExtensions{.pdf,.jpeg,.png,.jpg}

\usepackage[plain]{algorithm2e}
\usepackage{algorithmicx}
\usepackage{algpseudocode}
\usepackage{algcompatible}

\usepackage{booktabs}
\usepackage{multirow}
\usepackage[table]{xcolor}
\usepackage{siunitx}

\newcommand{\best}[1]{\cellcolor{green!15}\textbf{#1}}
\newcommand{\worst}[1]{\cellcolor{red!10}\textit{#1}}

\algblockdefx{Event}{EndEvent}[1]%
{\textbf{upon event }#1 \textbf{do }}%
{\textbf{}}

\algblockdefx{Function}{EndFunction}[1]%
{\textbf{Function }#1 }%
{\textbf{}}

\definecolor{mymaroon}{RGB}{128,0,0}

\usepackage[
  colorlinks=true,
  linkcolor=purple,
  citecolor=purple,
  urlcolor=purple,
  pdfauthor={},
  pdftitle={},
  pdfsubject={},
  pdfkeywords={},
  bookmarks=false,
]{hyperref}

\hypersetup{colorlinks=true,urlcolor=purple,citecolor=purple,urlcolor=purple}

\usepackage{pifont}





\usepackage{mathtools}
\newtagform{blue}{\color{BlueViolet}(}{)}

\DeclareUnicodeCharacter{0301}{\'{e}}

\usepackage[export]{adjustbox}

\begin{document}


\author{
\makebox[\textwidth][c]{%
\textbf{Antreas Ioannou}$^\dagger$ \quad
\textbf{Andreas Shiamishis}$^\dagger$ \quad
\textbf{Nora Hollenstein}$^\ddagger$ \quad
\textbf{Nezihe Merve Gürel}$^\dagger$
} \\[6pt]
\makebox[\textwidth][c]{%
$^\dagger$\textit{Delft University of Technology} \quad
$^\ddagger$\textit{University of Zurich}
} \\[6pt]
\makebox[\textwidth][c]{%
\{aioannou, ashiamishis, n.m.gurel\}@tudelft.nl \quad
nora.hollenstein@uzh.ch
}
}



\title{Evaluating the Limits of Large Language Models in \\Multilingual Legal Reasoning}
\date{}
\maketitle

\section*{Abstract}
In an era dominated by Large Language Models (LLMs), understanding their capabilities and limitations, especially in high-stakes fields like law, is crucial. While LLMs such as Meta’s LLaMA, OpenAI’s ChatGPT, Google’s Gemini, DeepSeek, and other emerging models are increasingly integrated into legal workflows, their performance in multilingual, jurisdictionally diverse, and adversarial contexts remains insufficiently explored. This work evaluates LLaMA and Gemini on multilingual legal and non-legal benchmarks, and assesses their adversarial robustness in legal tasks through character and word-level perturbations. We use an LLM-as-a-Judge approach for human-aligned evaluation. We moreover present an open-source, modular evaluation pipeline designed to support multilingual, task-diverse benchmarking of any combination of LLMs and datasets, with a particular focus on legal tasks, including classification, summarization, open questions, and general reasoning. Our findings confirm that legal tasks pose significant challenges for LLMs with accuracies often below 50\% on legal reasoning benchmarks such as LEXam, compared to over 70\% on general-purpose tasks like XNLI. In addition, while English generally yields more stable results, it does not always lead to higher accuracy. Prompt sensitivity and adversarial vulnerability is also shown to persist across languages. Finally, a correlation is found between the performance of a language and its syntactic similarity to English. We also observe that LLaMA is weaker than Gemini, with the latter showing an average advantage of about 24 percentage points across the same task. Despite improvements in newer LLMs, challenges remain in deploying them reliably for critical, multilingual legal applications.

\section{Introduction}
In the modern, LLM-dominated era, it is more important than ever to understand the capabilities and limitations of the black-box algorithms that accelerate human productivity across virtually every field~\cite{bommasaniOpportunitiesRisksFoundation2022a}. While the use of Large Language Models can be casual and insignificant at times, it is not rare for them to be deployed in critical, high-stakes scenarios and contexts where people's lives may be directly affected. One such area is the legal domain, where the use of LLMs for sensitive issues has become increasingly prominent~\cite{jackowskiFirstGlobalReport2023}.

Companies’ tendencies to maximize the productivity of their legal teams while minimizing expenses naturally lead them to adopt Large Language Models~\cite{joshiGenerativeAILegal2025, ogundeLegalLargeLanguage2025}, gradually increasing their reliance on technologies they may not fully understand. Moreover, in many cases the black-box nature of the LLMs make them less trusted within legal firms~\cite{reutersChatGPTGenerativeAi2023}. These situations make it necessary not only to evaluate these models on established legal benchmarks~\cite{chalkidisMultiEURLEXMultilingualMultilabel2021,aumillerEURLexSumMultiCrosslingual2022a, salaunEUROPALegalMultilingual2024a, drawzeski-etal-2021-corpus, fanLEXamBenchmarkingLegal2025}, but also to test them in multilingual settings to better understand how they process data and generate responses. It is only after thorough research that we should trust these models to make decisions worth millions of dollars, and potentially, thousands of lives.

Legal tasks present particular challenges for LLMs due to the specialized nature of legal language and reasoning. Legal texts are written in highly technical, domain-specific language, with complex sentence structures and terminology that diverges from everyday usage~\cite{mellinkoffLanguageLaw2004}. Such documents frequently contain long sentences, nested clauses, and precise formulations where even subtle shifts in wording can change the meaning dramatically~\cite{williamsTraditionChangeLegal2007}. Beyond linguistic complexity, legal reasoning requires contextual interpretation and sensitivity to jurisdictional variation~\cite{surdenArtificialIntelligenceLaw, chalkidis2022lexglue}. These characteristics make legal tasks more demanding than many general-purpose NLP benchmarks, and help explain why LLMs may underperform in this domain despite their broad capabilities~\cite{dahlLargeLegalFictions2024, huJHEvaluatingRobustness2025}.

Despite the growing use of LLMs in legal workflows, most existing evaluations focus on English-only tasks and do not reflect the multilingual, jurisdictionally diverse reality of legal practice. Additionally, the number of studies addressing the robustness of these models in the legal field under adversarial or ambiguous input, conditions that are common in real-world legal documents, remains significantly limited~\cite{dahlLargeLegalFictions2024, huJHEvaluatingRobustness2025}. As a result, there exists a significant gap in understanding how reliably these models perform across languages and legal contexts.

This work aims to address that gap. We hypothesize that LLMs perform significantly better in English than in other languages~\cite{zhang2023donttrustchatgptquestion}~\footnote{In this study we consider 15 different languages: Arabic, Bulgarian, German, Greek, English, Spanish, French, Italian, Maltese, Polish, Russian, Swedish, Thai, Turkish and Chinese}, both in terms of accuracy and stability. We also anticipate that their performance degrades under adversarial conditions~\cite{wang2024decodingtrustcomprehensiveassessmenttrustworthiness}, especially for legal tasks, revealing vulnerability in their reasoning processes. Moreover, we expect legal tasks to be particularly challenging for LLMs~\cite{chalkidis2022lexglue, dahlLargeLegalFictions2024}.

To test these hypotheses, we conduct a comprehensive evaluation of two state-of-the-art LLMs, Meta’s LLaMA and Google’s Gemini, across a suite of legal and general-purpose benchmarks~\cite{chalkidisMultiEURLEXMultilingualMultilabel2021,
aumillerEURLexSumMultiCrosslingual2022a,
salaunEUROPALegalMultilingual2024a,
drawzeski-etal-2021-corpus,
fanLEXamBenchmarkingLegal2025,
conneauXNLIEvaluatingCrosslingual2018,
artetxeCrosslingualTransferabilityMonolingual2020a}. Specifically, we evaluate LLaMA 3.1–8B and LLaMA 3.2–3B, as well as Gemini 1.5 Flash and the preview release of Gemini 2.5 Flash. We select these models to balance practical accessibility and architectural diversity. Our choice is also shaped by budgetary constraints, as both models provide free or low-cost access suitable for reproducible research. The versions chosen allow us to compare the models across different sizes and generations, highlighting trade-offs in performance, robustness, and multilingual generalization. The evaluation includes multilingual tasks such as legal classification, summarization, fairness prediction, and legal reasoning, as well as adversarial robustness tests. The datasets we use span a range of complexity and language coverage, enabling a fine-grained analysis of model performance in realistic legal settings.

Additionally, we augment traditional evaluation metrics with the use of an LLM-as-a-Judge framework, in which an LLM is tasked with evaluating the quality of model outputs~\cite{zhengJudgingLLMasaJudgeMTBench2023, liLLMsasJudgesComprehensiveSurvey2024, guSurveyLLMasaJudge2025}. This enables a more semantically informed, human-aligned assessment which is particularly valuable in tasks where correctness is context-sensitive or under-specified. By integrating this approach, we aim to move beyond surface-level metrics and better capture the nuanced reasoning required in legal domains.

We present the results of our evaluation across multilingual legal and general-purpose benchmarks, focusing on both accuracy and robustness. Our findings partially validate our hypotheses: English generally exhibits greater stability than other languages, as measured by variance and other stability metrics between different runs, and complex legal tasks indeed pose significant challenges for current LLMs. Moreover, a level of correlation is found between the performance of a language and its syntactic similarity to English. More specifically, there is a general upward trend in performance as a language becomes more syntactically similar to English. Under adversarial conditions, entropy between different runs tends to be lower for languages similar to English, while consistency is higher, highlighting the models’ ability to produce more stable and reliable predictions when the linguistic syntax aligns more closely with that of English. However, we also observe cases where English is outperformed by other languages in terms of accuracy, indicating that performance is not uniformly aligned with language resource availability.

Furthermore, our analysis reinforces the findings of previous studies~\cite{salechaLargeLanguageModels2024, scherrerEvaluatingMoralBeliefs2023}, that LLMs tend to default to more neutral or positive answers, particularly in tasks requiring nuanced judgment such as fairness classification. Moreover, when directly comparing Gemini and LLaMA, we observe that Gemini consistently outperforms LLaMA across the same evaluated task, with an average advantage of about 24 percentage points, highlighting Gemini’s stronger capabilities. We also highlight the substantial performance improvements and weaknesses observed in the newer Gemini model compared to its predecessor in multilingual legal contexts, suggesting that commercial LLMs are rapidly evolving but still face limitations in high-stakes, context-specific, multilingual applications.

\section{Related Work}
In this work, we evaluate LLMs at the intersection of multilingual understanding, legal-domain reasoning, and adversarial robustness. This section reviews the most relevant contributions in each domain and highlights prior work addressing their overlap.

Multilingual capabilities of LLMs is evaluated in various contexts. Zhang et al.~\cite{zhang2023donttrustchatgptquestion} show that models like ChatGPT struggle with translation-variant tasks, like pun detection. The MEGA benchmark~\cite{ahuja2023megamultilingualevaluationgenerative} evaluates generative LLMs on 70 languages across 16 datasets, exposing performance gaps in low-resource settings.

Cross-lingual evaluation has become a major focus in NLP. Benchmarks such as XNLI~\cite{conneauXNLIEvaluatingCrosslingual2018}, XTREME~\cite{hu2020xtreme}, and XGLUE~\cite{liang2020xglue} evaluate models across many languages and a variety of tasks, such as classification and named entity recognition. Manafi and Krishnaswamy~\cite{manafi2024crosslingual} further demonstrate that cross-lingual transfer depends on lexical overlap between languages. These studies underline the importance of multilingual evaluation and training, as model accuracy and biases vary substantially by language.

In the legal domain, Koreeda and Manning~\cite{koreeda2021contractnlidatasetdocumentlevelnatural} contribute ContractNLI, a document-level NLI dataset with evidence annotations, and introduce the Span NLI model that they conclude outperforms established transformer-based models. Domain-specific pretrained models like LEGAL-BERT~\cite{chalkidis2020legal} combined with an expanded grid search have improved performance on legal tasks. The LexGLUE benchmark~\cite{chalkidis2022lexglue} demonstrates that legal-domain pretraining consistently yields gains across classification, entailment, and summarization. ToS~\cite{drawzeski-etal-2021-corpus}  introduces the first annotated multilingual corpus of online Terms of Service documents for detecting potentially unfair clauses across four languages, and demonstrates a projection-based method for transferring English annotations to other languages using sentence embeddings. Kornilova and Eidelman~\cite{kornilova2019billsum} introduce BillSum, the first large-scale dataset for summarizing US legislative texts, and establish benchmark extractive summarization methods, showing that models trained on Congressional bills can effectively transfer to state-level legislation like California bills.

At the intersection of multilinguality and law, Aumiller et al.~\cite{aumillerEURLexSumMultiCrosslingual2022a} introduce EUR-Lex-Sum, a cross-lingual dataset for long-form summarization of EU legal texts in 24 languages. Chalkidis et al.~\cite{chalkidisMultiEURLEXMultilingualMultilabel2021} present MultiEURLEX for multilingual legal topic classification and show that simple adaptation techniques help multilingual models avoid forgetting other languages during fine-tuning, leading to better zero-shot performance, especially when there are many labels, and reveal that using time-based data splits is important due to changes in legal topics over time. EUROPA~\cite{salaunEUROPALegalMultilingual2024a} introduces a multilingual dataset for keyphrase generation in EU legal texts and presents experiments with multilingual models, highlighting challenges in cross-lingual performance and temporal generalization. LEXam~\cite{fanLEXamBenchmarkingLegal2025} introduces a multilingual benchmark for legal reasoning based on 340 law exams and evaluates LLMs across languages, underscoring significant performance gaps and the challenges of multilingual legal understanding.

Regarding adversarial robustness, Wang et al.~\cite{wang2024decodingtrustcomprehensiveassessmenttrustworthiness} provide a comprehensive benchmark that evaluates GPT-3.5 and GPT-4 in key dimensions of trustworthiness, including toxicity, bias, robustness, privacy and fairness, revealing that while GPT-4 is generally more reliable, it is also more vulnerable to adversarial requests due to its stronger follow-up behavior of instructions. Upadhayay and Behzadan~\cite{upadhayay2024sandwich} propose a multilingual prompt injection (Sandwich attack) that bypasses content filters. Manafi and Krishnaswamy~\cite{manafi2024crosslingual} find that cross-lingual training offers slight robustness improvements. Raj and Devi~\cite{raj2023legalrobustness} reveal that legal-domain models are especially vulnerable to semantic and syntactic perturbations.

However, as Dahl et al.~\cite{dahlLargeLegalFictions2024} observe, while failures of LLMs in legal settings have been widely discussed, systematic evaluations of adversarial robustness in legal contexts remain limited. Hu et al.~\cite{huJHEvaluatingRobustness2025} further demonstrate that even subtle semantic or syntactic variations in legal text can significantly disrupt model predictions, emphasizing the fragility of LLMs in real-world legal scenarios.

The intersection of multilingual legal capabilities and adversarial robustness remains critically understudied. This gap is echoed in recent studies~\cite{dahlLargeLegalFictions2024, huJHEvaluatingRobustness2025}, which call for more comprehensive evaluation frameworks. With our research we aim to cover this gap in existing work and provide a thorough analysis on the capabilities and limitations of LLMs in these domains, that jointly considers multilingual, legal, and adversarial dimensions.

\section{Preliminaries}
\label{setup}
This section outlines the Large Language Models we evaluate, the datasets we utilize for benchmarking, and the specific evaluation metrics we apply to measure performance for each dataset.

\subsection{Models}
\textbf{Gemini 1.5 Flash}, developed by Google DeepMind and released in 2024, is a lighter variant of the Gemini 1.5 family. Its inclusion in this study is motivated by its increasing adoption in enterprise and public-facing applications where both efficiency and linguistic robustness are critical. It represents the current state of the art in commercial LLM deployment and provides a relevant baseline for legal and multilingual evaluation.

\textbf{Gemini 2.5 Flash (Preview 05/20)}, also developed by Google DeepMind, is an early-access version of the next-generation Gemini 2.5 series. Released in May 2025 as a developer preview, this model builds on the Gemini 1.5 Flash architecture with expected improvements in reasoning accuracy, instruction-following, and response stability. We include this preview release to examine forward-looking capabilities in emerging commercial LLMs and to assess incremental gains over its 1.5 predecessor in complex domain-specific tasks.

\textbf{LLaMA 3.1–8B}~\cite{touvron2023llama}, developed by Meta and released in 2024, is an open-weight large language model designed to balance performance and efficiency across a wide range of tasks. As part of the LLaMA 3 family, the 8B variant offers strong general-purpose reasoning, competitive multilingual capabilities, and support for extended context windows, making it well-suited for both research and practical deployment in resource-constrained environments. Its relatively modest parameter size enables faster inference and easier fine-tuning compared to larger models, while still achieving robust performance on benchmarks spanning legal, technical, and everyday language domains. This model is included in our study as a representative open-weight baseline, reflecting ongoing advancements in accessible, high-quality LLMs for multilingual and domain-sensitive applications. We use the LLaMA 3.1–8B model served via Ollama, an open-source tool that enables local deployment of large language models with minimal setup.

\textbf{LLaMA 3.2–3B} is a lightweight variant of Meta’s LLaMA 3 family, offering a compact yet capable open-weight large language model released in 2024. Despite its smaller parameter size, the 3B model demonstrates solid performance on a range of tasks, including multilingual understanding and domain-specific reasoning, making it particularly suitable for experimentation and deployment in low-resource or latency-sensitive settings. Its efficiency enables rapid inference and fine-tuning on consumer-grade hardware, while still benefiting from the architectural and training improvements introduced in the LLaMA 3 series. In our study, LLaMA 3.2–3B serves as a baseline to assess the trade-off between model size and performance in multilingual and legal applications.

The models are selected based on budget constraints and diversity in model architecture. Due to our limited budget we resort to LLMs that offer free access to their services. This can be important for agents who aim to utilize Language Models but also operate under financial limitations.
Furthermore, we choose the two models, Gemini and LLaMA, in an attempt to cover two distinct classes of LLMs, a multi-modal, enhanced one, in Gemini, and a vanilla one in LLaMA.

For Gemini, we access the models via the official Google API~\footnote{\url{https://ai.google.dev/gemini-api/docs}}
, using the endpoints corresponding to Gemini 1.5 Flash (2024 release) and Gemini 2.5 Flash (Preview 05/20). For LLaMA, we rely on the Ollama framework~\footnote{\url{https://ollama.com/library}}
to serve the LLaMA 3.1–8B~\footnote{\url{https://ollama.com/library/llama3.1:8b}} and LLaMA 3.2–3B~\footnote{\url{https://ollama.com/library/llama3.2:3b}} models.

\subsection{Evaluation Metrics}
\label{evaluation_metrics}

To capture the diverse requirements of the tasks considered, we employ a broad set of evaluation metrics spanning four categories: classification metrics (Accuracy, Precision, Recall, F1, mRP) to assess correctness in structured prediction tasks; text generation metrics (ROUGE, BLEU, METEOR, Cosine Similarity) to evaluate the quality of summaries and free-form outputs; robustness and reliability metrics (variance measures, consistency, entropy, Gini Index, confidence margin) to quantify stability under repeated runs and adversarial conditions; and LLM-as-judge scores to capture quality judgments beyond surface similarity. This combination ensures that our evaluation does not privilege a single aspect of performance but instead provides a balanced view of predictive accuracy, linguistic quality, semantic faithfulness, and robustness.

\begin{table}[H]
\centering
\small
\resizebox{\textwidth}{!}{%
\begin{tabular}{|p{3.5cm}|p{5.5cm}|p{5.5cm}|p{4.5cm}|}
\hline
\textbf{Metric} & \textbf{Formula} & \textbf{Explanation} & \textbf{Description} \\
\hline
Accuracy & \makecell[l]{\rule{0pt}{3.5ex}$\frac{1}{N} \sum_{i=1}^{N} \mathbf{1}(y_i = \hat{y}_i)$} & $N$ = number of examples, $y_i$ = true label, $\hat{y}_i$ = predicted label, $\mathbf{1}$ = indicator function & Proportion of correct predictions among all predictions \\
\hline
Precision & \makecell[l]{\rule{0pt}{3.5ex}$\frac{\text{TP}}{\text{TP} + \text{FP}}$} & $\text{TP}$ = true positives, $\text{FP}$ = false positives & Proportion of predicted positives that are actually correct \\
\hline
Recall & \makecell[l]{\rule{0pt}{3.5ex}$\frac{\text{TP}}{\text{TP} + \text{FN}}$} & $\text{TP}$ = true positives, $\text{FN}$ = false negatives & Proportion of actual positives that are correctly predicted \\
\hline
F1 Score & \makecell[l]{\rule{0pt}{3.5ex}$2 \cdot \frac{\text{Precision} \cdot \text{Recall}}{\text{Precision} + \text{Recall}}$} & Uses values from Precision and Recall & Harmonic mean of Precision and Recall \\
\hline
Mean R-Precision (mRP) & \makecell[l]{\rule{0pt}{3.5ex}$\frac{1}{N} \sum_{i=1}^{N} \frac{|\text{Top-}k_i(\hat{y}_i) \cap y_i|}{k_i}$} & $k_i$ = \# of true labels, $\hat{y}_i$ = top-$k$ predicted labels, $y_i$ = true label set & Mean precision at rank $k$, where $k$ = number of true labels \\
\hline
ROUGE-1 & 
\makecell[l]{\rule{0pt}{3.5ex}
$\text{ROUGE-1} = \frac{2PR}{P + R}$ \\
$P = \frac{|U_{\text{ref}} \cap U_{\text{gen}}|}{|U_{\text{gen}}|}$, 
$R = \frac{|U_{\text{ref}} \cap U_{\text{gen}}|}{|U_{\text{ref}}|}$} & 
$U_{\text{ref}}$, $U_{\text{gen}}$ = unigrams in reference and generated text & 
F1 score over unigram overlap. Captures lexical similarity. \\
\hline
ROUGE-2 & 
\makecell[l]{\rule{0pt}{3.5ex}
$\text{ROUGE-2} = \frac{2PR}{P + R}$ \\
$P = \frac{|B_{\text{ref}} \cap B_{\text{gen}}|}{|B_{\text{gen}}|}$, 
$R = \frac{|B_{\text{ref}} \cap B_{\text{gen}}|}{|B_{\text{ref}}|}$} & 
$B_{\text{ref}}$, $B_{\text{gen}}$ = bigrams in reference and generated text & 
F1 score over bigram overlap. Reflects fluency and phrase structure. \\
\hline
ROUGE-L & 
\makecell[l]{\rule{0pt}{3.5ex}
$\text{ROUGE-L} = \frac{2PR}{P + R}$ \\
$P = \frac{\text{LCS}(\text{ref}, \text{gen})}{|\text{gen}|}$, \quad 
$R = \frac{\text{LCS}(\text{ref}, \text{gen})}{|\text{ref}|}$} & 
LCS = Longest Common Subsequence between reference and generated text & 
F1 score based on LCS over flat token sequences. Captures token-level order and structure. \\
\hline
ROUGE-L Sum & 
\makecell[l]{\rule{0pt}{3.5ex}
$\text{ROUGE-Lsum} = \frac{2PR}{P + R}$ \\
$P = \frac{\text{LCS}(S_{\text{ref}}, S_{\text{gen}})}{|S_{\text{gen}}|}$, \quad 
$R = \frac{\text{LCS}(S_{\text{ref}}, S_{\text{gen}})}{|S_{\text{ref}}|}$} & 
$S_{\text{ref}}$, $S_{\text{gen}}$ = sentence-tokenized versions of reference and generated summaries & 
F1 score based on LCS after sentence-level tokenization. Optimized for evaluating multi-sentence summaries. \\
\hline
BLEU & - & - & Precision of n-gram overlaps between generated and reference texts \\
\hline
METEOR & - & - & Considers synonymy, stemming, and recall for n-gram overlaps \\
\hline
Cosine Similarity & \makecell[l]{\rule{0pt}{3.5ex}$\frac{\vec{a} \cdot \vec{b}}{\|\vec{a}\| \|\vec{b}\|}$} & $\vec{a}, \vec{b}$ = vector representations of generated and reference texts & Cosine of the angle between two vectors; measures semantic similarity \\
\hline
Per-sample Performance Variance & \makecell[l]{\rule{0pt}{3.5ex}$\frac{1}{N} \sum_{i=1}^{N} (x_i - \bar{x})^2$} & $x_i$ = score for example $i$, $\bar{x}$ = mean score, $N$ = number of examples & Variance of a continuous performance metric across samples \\
\hline
Inter-run Prediction Variance & 
\makecell[l]{\rule{0pt}{3.5ex}$\frac{1}{N} \sum_{i=1}^{N} \text{Var}(X_i)$} & 
$X_i$ = set of predictions for $i$-th example across $n$ runs & 
Prediction variance across multiple runs \\
\hline
Shannon Entropy & \makecell[l]{\rule{0pt}{3.5ex}$- \sum_{j=1}^{C} p_j \log_2 p_j$} & $p_j$ = probability of class $j$, $C$ = number of classes & Measures prediction uncertainty using class probabilities \\
\hline
Prediction Consistency & \makecell[l]{\rule{0pt}{3.5ex}$\frac{1}{N} \sum_{i=1}^{N} \frac{\max_{c \in \mathcal{C}} \text{Count}_i(c)}{n}$} & $\text{Count}_i(c)$ = count of class $c$ predicted for $i$-th example over $n$ runs & Frequency of most common prediction across runs \\
\hline
LLM Judge Score & \makecell[l]{\rule{0pt}{3.5ex}$\frac{1}{N} \sum_{i=1}^{N} s_i$} & $s_i$ = LLM score for example $i$, $N$ = number of examples & Average LLM-assigned score for model outputs \\
\hline
Gini Index & $G = 1 - \sum_{i=1}^{C} p_i^2$ & $p_j$ = probability of class $j$, $C$ = number of classes. & Measures uncertainty of the prediction distribution. Lower values indicate higher confidence.  \\
\hline
Confidence Margin & $CM = p_{\text{top}} - p_{\text{second}}$ & $p_{\text{top}}$ is the highest softmax probability and $p_{\text{second}}$ is the second-highest.
 & Difference between the highest and second-highest predicted probabilities. Higher values indicate more confident predictions. \\
\hline
\end{tabular}
}
\caption{Summary of evaluation metrics with formulas, explanations, and descriptions.}
\label{tab:evaluation_metrics}
\end{table}

\subsection{Datasets}
\label{datasets}
For our evaluation, we consider seven datasets, five of which belong to the legal domain. We select these seven datasets to ensure a comprehensive and reliable evaluation across both legal-domain and general-purpose NLP tasks, with a focus on multilingual and task-diverse benchmarks. All datasets are sourced from the Hugging Face \texttt{datasets} library~\cite{lhoestDatasetsCommunityLibrary2021}, ensuring consistent loading, preprocessing, and integration across experiments. The five legal datasets (MultiEURLEX, Eur-Lex-Sum, EUROPA, ToS, and LEXam) are chosen as the most reliable benchmarks available in Hugging Face for their respective tasks, including multilabel classification, summarization, keyphrase generation, fairness assessment, and legal reasoning. Together, they span the major categories of legal NLP applications, enabling a broad and representative assessment of model performance within the legal domain. All five are multilingual, supporting evaluation of cross-lingual generalization. To establish a baseline for general-language capabilities, we include XNLI and XQuAD, which are standard benchmarks for inference and extractive question answering. This allows us to distinguish domain-specific performance from general linguistic competence. This selection ensures a rigorous, fair, and scalable evaluation framework.

\subsubsection{Legal} 
\textbf{MultiEURLEX}~\cite{chalkidisMultiEURLEXMultilingualMultilabel2021} is a multilingual dataset for topic-based multilabel classification of legal documents. It comprises 65,000 European Union (EU) legislative acts that are officially translated into 23 languages. Each document is annotated with multiple labels drawn from the EUROVOC Level 3 taxonomy, which includes 567 fine-grained legal and policy topics. The taxonomy is hierarchical: Level 1 covers broad domains (e.g., \textit{Politics}, \textit{Economics}), Level 2 introduces intermediate groupings, and Level 3 provides specific subcategories used in practical document tagging. For evaluation, we use standard multilabel metrics including \textbf{precision}, \textbf{recall}, and \textbf{F1 score}. Additionally, we report \textbf{mean R-Precision (mRP)}.
\\\\
\textbf{Eur-Lex-Sum}~\cite{aumillerEURLexSumMultiCrosslingual2022a} is a multilingual dataset for summarization based on European Union legislative texts. The dataset comprises legal documents officially translated into 24 EU languages and paired with human-written summaries. The summarization task involves generating a condensed representation of each legal document. Model performance on this task is evaluated using the ROUGE metric family, including \textbf{ROUGE-1}, \textbf{ROUGE-2}, \textbf{ROUGE-L}, and \textbf{ROUGE-L Sum}, which collectively assess content overlap, fluency, and structural similarity between generated and reference summaries. We also report \textbf{cosine similarity} as an alternative metric.
\\\\
\textbf{EUROPA}~\cite{salaunEUROPALegalMultilingual2024a} is a multilingual dataset designed for training and evaluating keyphrase generation models in the legal domain. It comprises legal judgments from the Court of Justice of the European Union (CJEU), officially translated into all 24 EU official languages. Each document is paired with a set of reference keyphrases, manually curated to encapsulate the core legal concepts. The dataset is available in various splits; in this study, we utilize the random split, which partitions the data randomly into training, validation, and test sets. Model performance is evaluated using the \textbf{LLM Judge Score} and \textbf{per-sample performance variance}.
\\\\
\textbf{Online Terms of Service (ToS)}~\cite{drawzeski-etal-2021-corpus} is a multilingual dataset focused on fairness classification of legal clauses in consumer contracts. It contains sentences extracted from real-world online terms of service documents, annotated with one of three fairness labels: \textit{clearly\_fair}, \textit{potentially\_unfair}, or \textit{clearly\_unfair}. The dataset is intended for evaluating models on their ability to distinguish nuanced legal judgments related to consumer protection. For evaluation, we use the standard \textbf{accuracy}, as well as a \textbf{custom penalty rating} that quantifies the severity of misclassifications: assigning a score of 0 for correct predictions, 0.5 for near misses (e.g., predicting "potentially unfair" instead of "clearly unfair"), and 1.0 for severe misjudgments (e.g., predicting "clearly fair" for a "clearly unfair" clause).

\textbf{LEXam}~\cite{fanLEXamBenchmarkingLegal2025} is a multilingual benchmark comprising 4,886 real-world law exam questions from 340 exams across 116 law school courses, spanning a wide range of subjects and degree levels. It is designed to evaluate legal reasoning and exam-style comprehension in both \textit{open-ended} and \textit{multiple-choice} formats. The dataset includes 2,841 open questions with expert-written reference answers and reasoning guidance, as well as 2,045 multiple-choice questions. Questions span diverse legal domains, including civil, criminal, constitutional, and international law, and are available in both English and German. The benchmark emphasizes structured legal reasoning, requiring models to identify legal issues, recall legal norms, and apply them in context. In this study, we evaluate model performance on both the open-ended and the four-choice multiple-choice subsets. For evaluating the multiple-choice questions, we use \textbf{accuracy} across data points. For the open-ended questions, we employ the \textbf{LLM Judge Score} along with its \textbf{per-sample performance variance}, as these metrics are better suited to capture legal correctness and reasoning quality than those based solely on lexical or semantic overlap.

\subsubsection{Non-Legal}

\textbf{XNLI}~\cite{conneauXNLIEvaluatingCrosslingual2018} is a multilingual natural language inference dataset designed to evaluate cross-lingual understanding. It consists of sentence pairs in 15 languages derived from the MultiNLI corpus, where the task is to classify the relationship between a premise and a hypothesis as either \textit{entailment}, \textit{contradiction}, or \textit{neutral}. This task evaluates the model's ability to reason about semantic relationships across languages. Model performance on this dataset is evaluated using \textbf{accuracy}, as well as robustness metrics including \textbf{inter-run prediction variance}, \textbf{Shannon entropy}, and \textbf{prediction consistency} across multiple runs.
\\\\
\textbf{XQuAD}~\cite{artetxeCrosslingualTransferabilityMonolingual2020a} is a multilingual dataset designed for extractive question answering, where the model is required to generate an answer based on a given context passage. The dataset supports 11 languages, including English, Spanish, German, Russian, Arabic, and Hindi, enabling evaluation of cross-lingual generalization and language-specific performance. Each instance includes a natural language question, a reference answer, and a context passage from which the answer can be inferred. The task assesses the model’s ability to comprehend the passage and generate an accurate and contextually relevant response. Model performance is evaluated using \textbf{BLEU}, \textbf{METEOR}, and \textbf{Cosine Similarity}, which together assess both lexical and semantic overlap between the generated and reference answers. Additionally, to address potential bias in embedding-based metrics (e.g., favoring English), we also include the \textbf{LLM Judge Score} and \textbf{per-sample performance variance} as complementary evaluation metrics.

\section{Methodology}
\begin{figure}[t!]
\centering
\includegraphics[width=0.65\textwidth]{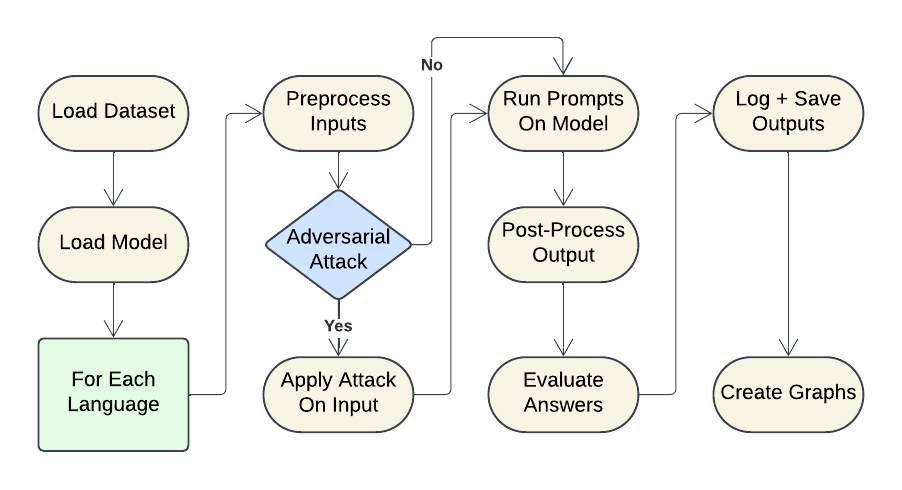}
\caption{Overview of the multilingual LLM evaluation pipeline.}
\label{fig:pipeline}
\end{figure}

An important contribution of this work is the development of a well-structured and extensible evaluation pipeline that supports any combination of models and datasets.\footnote{Code available at: \url{https://github.com/RobustML-Lab/Legal-Multilingual-Evaluation-of-LLMs}} In this section, we describe the evaluation process and the overall architecture of the pipeline. A graphical overview is presented in Figure~\ref{fig:pipeline}.

The pipeline accommodates both classification and text generation tasks, and is compatible with multilingual datasets. It is designed with repeatability, scalability, and modularity in mind, enabling seamless integration with additional features, including LLM-based evaluators.

In each experiment, the system iterates over all languages available in a given dataset, or a specified subset, and generates predictions accordingly. These predictions are then post-processed: for instance, labels are extracted in classification tasks, enabling comparison between the model outputs and the ground truth.

The evaluation process adapts to the input type and the target metrics. For classification, we compute standard statistical metrics such as accuracy, precision, recall, and F1 score. For generative tasks, we employ different metrics such as cosine similarity, ROUGE scores, and LLM judge scores, according to the task, to evaluate semantic and lexical alignment.

The pipeline also supports optional adversarial robustness testing. When enabled, inputs are perturbed using a configurable attack module prior to inference. Supported attacks include character-level manipulations and context-aware word substitutions. Both original and perturbed inputs are retained for comparative evaluation.

Outputs from each run, including raw model generations, processed predictions, gold references, and evaluation scores, are stored in a structured format on disk, ensuring reproducibility and traceability. When necessary, model responses are post-processed to extract the expected output format (e.g., mapping a free-text answer to a classification label), with the processing steps aligned to the prompting strategy. This ensures that the stored outputs are consistent and directly usable for downstream analysis. After each run, performance metrics are aggregated per language (e.g., accuracy, F1), and additional tools allow for the inspection of the top-$N$ outputs or detailed per-sample results.

The codebase follows an object-oriented paradigm and is intentionally modular. It is model- and dataset-agnostic, relying on pluggable \texttt{Data} and \texttt{Model} components to manage input preprocessing and LLM querying. In practice, the datasets we use are loaded via the Hugging Face \texttt{datasets} library, but the design does not restrict this: adding a new dataset only requires implementing a child class with a \texttt{get\_data()} method, which could just as easily read from CSV files, custom corpora, or other APIs. Similarly, new models can be integrated by subclassing \texttt{Model} and implementing the prediction interface. This design makes it straightforward to extend the pipeline to new datasets and models with minimal effort.

\section{LLM Judge}
\label{llm_judge}
To complement standard evaluation metrics, we include an LLM-as-a-Judge (LLM Judge)~\cite{zhengJudgingLLMasaJudgeMTBench2023}, a large language model-based evaluation mechanism designed to provide semantically grounded assessments across tasks and languages. Recent studies~\cite{liLLMsasJudgesComprehensiveSurvey2024, guSurveyLLMasaJudge2025} have shown that LLM-based evaluation can offer deeper insights and better alignment with human judgment compared to traditional metrics.

The LLM Judge is implemented as a configurable module that can be enabled for any dataset\footnote{Our implementation of the LLM Judge follows practical guidelines outlined by Hugging Face's LLM Judge Cookbook, which offers best practices for prompt construction, score parsing, and evaluation setup. Available at: \url{https://huggingface.co/learn/cookbook/llm_judge}}. When activated, it replaces or augments traditional metric-based evaluation by generating dynamic prompts that are passed to a language model (\textbf{Gemini 2.0 Flash}). These prompts are constructed from the model's predictions and corresponding reference outputs and are designed to elicit a numerical quality score ranging from 1 to 5.

During evaluation, the LLM is tasked with assessing aspects such as semantic similarity, correctness of label extraction, or keyphrase relevance. The output from the LLM is parsed and aggregated to compute an \textbf{LLM Judge Score}, which reflects the average quality or correctness of the model's responses as judged by the LLM.

The rationale for using the LLM Judge is threefold:
\begin{itemize}
    \item \textbf{Semantic awareness:} Traditional metrics often rely on surface-level overlap, which may not accurately capture semantic equivalence, especially in multilingual or generative settings.
    \item \textbf{Robustness to format variation:} LLMs can interpret loosely structured outputs (e.g., label names or free-form summaries), enabling more flexible response formats from the evaluated models.
    \item \textbf{Language adaptability:} While prompts are written in English to maintain consistency and avoid introducing linguistic bias, the content being evaluated (e.g., input text, model outputs, and reference answers) is presented in the target language. This ensures that the LLM judge assesses performance within the intended linguistic context without being influenced by prompt translation quality.

\end{itemize}

The LLM Judge therefore serves as a more human-aligned evaluator, particularly effective in tasks where correctness is subjective, under-specified, or context-dependent.

\section{Adversarial Attacks}
Adversarial attacks have become a key technique for evaluating the robustness of language models when exposed to small but strategically crafted input perturbations. Previous work~\cite{liTextBuggerGeneratingAdversarial2019, wang2024decodingtrustcomprehensiveassessmenttrustworthiness} demonstrates that even high-performing models are often brittle: minor modifications, such as introducing typos or replacing words with synonyms, can significantly alter model behavior, sometimes yielding incorrect or even dangerous outputs.

While much of the literature focuses on attacking prompts or crafting jailbreaking strategies, our approach considers a more grounded threat model. Specifically, we apply adversarial attacks not to the model prompt but to the input data itself. This setup reflects real-world scenarios in legal and regulatory environments where the prompts (e.g., legal queries, standard questions) are fixed, but the input documents or user queries may be noisy, misformatted, or adversarial in nature.

After perturbing the input data, we pass it to the language models using the original, unmodified evaluation prompts. The models are then evaluated under the same conditions and metrics used in the clean data setting, allowing for a direct assessment of input-level robustness.

We implement and evaluate two adversarial attack techniques using the \texttt{nlpaug} \footnote{\url{https://github.com/makcedward/nlpaug}} Python library:

\begin{itemize}
    \item \textbf{Random Character Insertion (Typographic Noise):} This attack introduces noise at the character level by randomly inserting characters into input words. It simulates natural errors such as typos or Optical Character Recognition (OCR) \footnote{OCR (Optical Character Recognition) refers to the process of converting scanned images of text into machine-readable text. It is prone to introducing errors when characters are misread or misaligned, especially in low-quality scans.} noise and probes the model's resilience to surface-level disruptions. Despite being syntactically minor, these perturbations can significantly affect model comprehension.

    \item \textbf{Contextual Word Substitution (BERT-based):} Using pre-trained contextual embeddings, this method replaces words with semantically similar alternatives while maintaining grammatical and contextual coherence. This type of substitution poses a deeper challenge for models, testing their ability to preserve meaning in the face of lexical variation.
\end{itemize}

These adversarial strategies allow us to systematically probe different facets of model sensitivity: syntactic robustness through typographic noise and semantic robustness through paraphrasing. In our experiments, we perturb either 30\% of the tokens, following the default configuration of the \texttt{nlpaug} library, or 15\% of the tokens in cases where the default setting introduces excessive distortion for a given dataset. The choice of perturbation rate is thus consistent across datasets but adjusted when necessary to maintain meaningful input semantics. Results are presented in Section~\ref{evaluation}.

\section{Evaluation}
\label{evaluation}
The evaluation is performed using four different models across five legal and two non-legal datasets, with model coverage varying by dataset. Graphs include abbreviations of the languages using ISO 639-1 codes. Table~\ref{tab:lang_abbr} in Appendix \ref{appendix:abbreviations} provides the corresponding language mappings.

\subsection{Baseline Evaluation}
\label{baseline_evaluation}

We first present overall evaluation results and a graphic showcasing the aggregated results, before turning to detailed per-dataset analyses. In Table~\ref{tab:overview_clean_sections} we present a summary of the baseline evaluation results across all datasets and models. Since the tasks rely on different evaluation metrics, the values are not directly comparable across tasks; a detailed breakdown of each metric and dataset is provided in the following sections. For each dataset we present in the table what we consider the most informative metric to be from its respective evaluation. The purpose of this table is to give the reader a compact overview and to facilitate comparison within the same task, while highlighting the range of languages and models included in our study.

\begin{table*}[t]
\centering
\scriptsize
\resizebox{\textwidth}{!}{
\begin{tabular}{l *{15}{c}}
\toprule
Dataset (Metric) &
en & bg & el & fr & mt & sv & de & it & pl & es & th & ar & zh & ru & tr \\
\midrule
\multicolumn{16}{l}{\textbf{Gemini 1.5 Flash}} \\
MultiEURLEX (F1)
& \best{0.21} & \worst{0.18} & 0.19 & \best{0.21} & 0.20 & \best{0.21} & — & — & — & — & — & — & — & — & — \\
EUROPA (LLM Score)
& 3.77 & 3.76 & \worst{3.32} & 3.50 & \best{3.79} & 3.70 & — & — & — & — & — & — & — & — & — \\
ToS - Highly Assertive Prompt (Accuracy)
& \best{0.66} & — & — & — & — & — & 0.63 & \worst{0.61} & 0.65 & — & — & — & — & — & — \\
LEXam\textendash MC (Accuracy)
& \best{0.48} & — & — & — & — & — & \worst{0.40} & — & — & — & — & — & — & — & — \\
LEXam\textendash Open (LLM Score)
& \best{3.71} & — & — & — & — & — & \worst{2.77} & — & — & — & — & — & — & — & — \\
XNLI (Accuracy)
& \best{0.73} & 0.67 & 0.70 & 0.69 & — & — & — & — & — & \best{0.73} & \worst{0.56} & — & — & — & — \\
XQuAD (LLM Score)
& \best{4.94} & 4.73 & — & — & — & — & — & — & — & — & 4.74 & 4.67 & 4.68 & 4.78 & \worst{4.61} \\
\midrule
\multicolumn{16}{l}{\textbf{Gemini 2.5 Flash}} \\
LEXam\textendash MC (Accuracy)
& \best{0.74} & — & — & — & — & — & \worst{0.51} & — & — & — & — & — & — & — & — \\
LEXam\textendash Open (LLM Score)
& \best{4.14} & — & — & — & — & — & \worst{3.62} & — & — & — & — & — & — & — & — \\
\midrule
\multicolumn{16}{l}{\textbf{LLaMA 3.1\textemdash 8B}} \\
Eur\textendash Lex\textendash Sum (ROUGE-L Sum)
& 0.18 & \worst{0.05} & \worst{0.05} & \best{0.20} & — & — & 0.16 & — & 0.11 & — & — & — & — & — & — \\
\midrule
\multicolumn{16}{l}{\textbf{LLaMA 3.2\textemdash 3B}} \\
XNLI (Accuracy)      
& \best{0.52} & 0.47 & 0.45 & 0.43 & — & — & — & — & — & 0.41 & \worst{0.34} & — & — & — & — \\
\bottomrule
\end{tabular}
}
\caption{Summary of results across all datasets and models. Since different tasks use different evaluation metrics, values are only directly comparable within the same task. “—” indicates either that the language is not supported by the dataset or was not evaluated. Best and worst scores per row are highlighted.}
\label{tab:overview_clean_sections}
\end{table*}

In addition, in order to provide a graphic general view of multilingual performance, we adopt two complementary normalization strategies:  
(1) an aggregated per-language score across all models and datasets, and  
(2) per-dataset heatmaps that highlight relative performance differences within each dataset.  

From each evaluation, we report only one primary metric. For MultiEURLEX, we use the F1 score, as it balances precision and recall in the multilabel classification setting and is the standard metric for this task. For Eur-Lex-Sum, we adopt the ROUGE-L Sum score, which best captures the linguistic meaning of responses among the metrics used for this task. For the ToS dataset, we report accuracy under the highly assertive prompt configuration (further explained in Subsection~\ref{tos_section}), since this setup best reflects realistic usage where the evaluator can optimize the prompt for performance. For the LEXam benchmark, we include results for both the multiple-choice and the open-ended questions, as they capture complementary aspects of legal reasoning. For the remaining datasets, we follow their respective default evaluation metrics introduced in the corresponding subsections: EUROPA - LLM Score, LEXam (MC) - Accuracy, LEXam (Open) - LLM Score, XNLI - Accuracy, and XQuAD - LLM Score.

\paragraph{Aggregated scores.}  
Since each dataset uses a different metric scale (e.g., accuracy, F1, BLEU, average score), raw values are not directly comparable. 
To make them commensurate, we normalize per dataset using min-max scaling, average across models when more than one is available for the same dataset, and then aggregate across datasets.  

Formally, for model $m$, let $s_{d,l}^m$ denote the raw score of language $l$ on dataset $d$. 
We first normalize within each dataset:  

\[
\tilde{s}_{d,l}^m = \frac{s_{d,l}^m - \min_{l'} s_{d,l'}^m}{\max_{l'} s_{d,l'}^m - \min_{l'} s_{d,l'}^m}.
\]

When multiple models are available for dataset $d$, we average their normalized scores:  

\[
\hat{s}_{d,l} = \frac{1}{|M_d|} \sum_{m \in M_d} \tilde{s}_{d,l}^m,
\]

where $M_d$ is the set of models evaluated on $d$.  
Finally, we compute the aggregated score for each language as:  

\[
\bar{s}_{l} = \frac{1}{|D_l|} \sum_{d \in D_l} \hat{s}_{d,l},
\]

where $D_l$ is the set of datasets in which language $l$ was evaluated.  
The resulting $\bar{s}_{l}$ provides a normalized, task-independent indicator of a language’s relative performance, enabling direct comparison across languages despite differing metric scales.

\paragraph{Per-dataset heatmaps.}  
For dataset-level visualizations, we instead apply $z$-score normalization. This avoids distortions from min--max scaling in datasets with very few languages (e.g., two), where one language would be forced to $0$ and the other to $1$.  
Given the raw scores $s_{d,l}^m$, we compute  

\[
z_{d,l}^m = \frac{s_{d,l}^m - \mu_d^m}{\sigma_d^m},
\]

where $\mu_d^m$ and $\sigma_d^m$ are the mean and standard deviation of scores across all languages in dataset $d$ for model $m$.  
To make results comparable across datasets, we clip $z_{d,l}^m$ to a fixed range $[-c,c]$ (with $c=2$ in our experiments) and rescale into $[0,1]$:  

\[
\tilde{z}_{d,l}^m = \frac{\min(\max(z_{d,l}^m, -c), c) + c}{2c}.
\]

This preserves relative differences within datasets, while preventing extreme values from dominating. 

\begin{figure}[H]
    \centering
    \includegraphics[width=\linewidth]{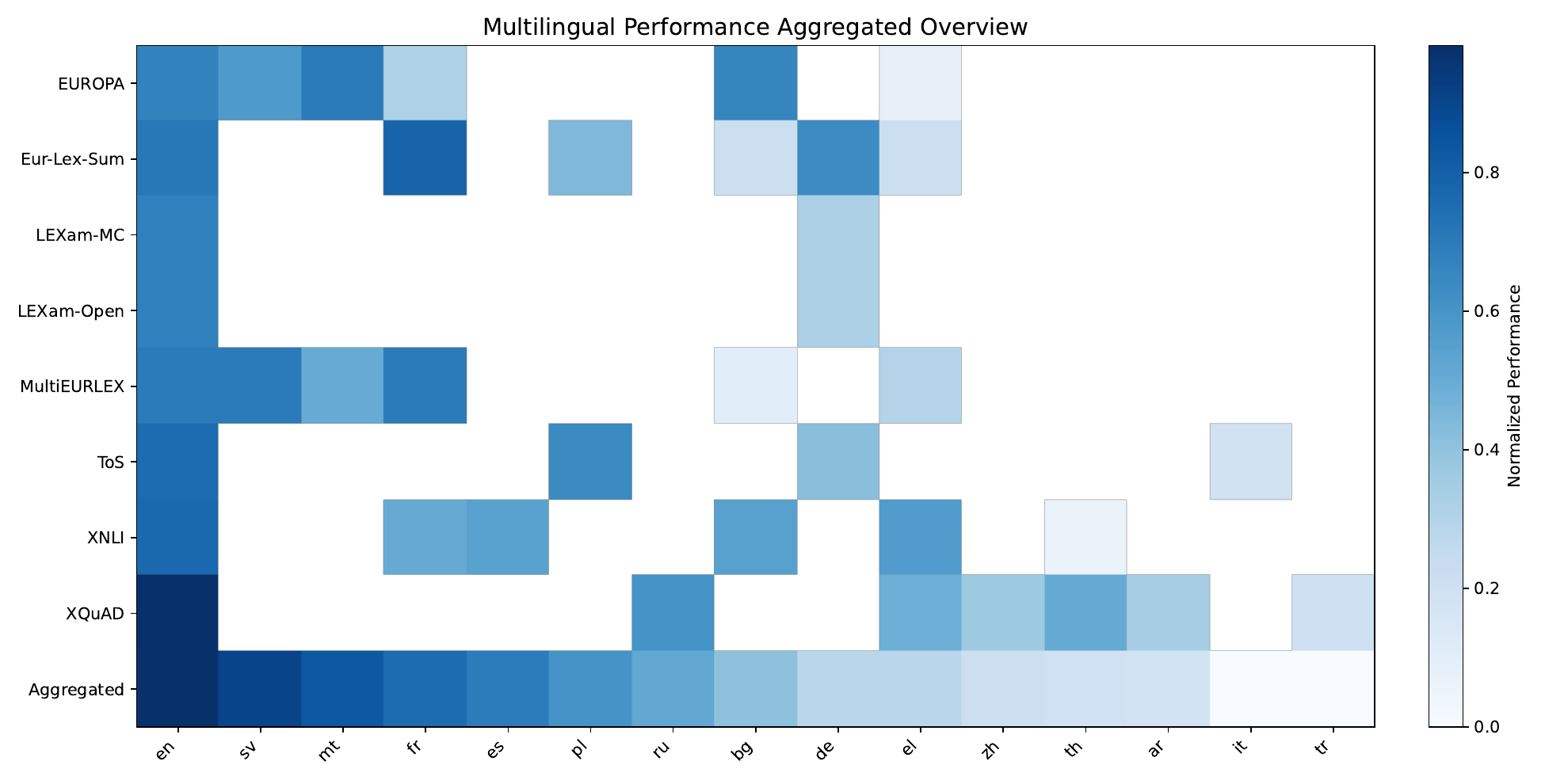}
    \caption{Overview of multilingual performance across datasets and models. 
    Rows correspond to datasets (normalized with $z$-scores) and the aggregated score (min-max normalization), 
    while columns correspond to languages sorted by their aggregated score. 
    Darker colors indicate stronger relative performance.}
    \label{fig:combined_heatmap}
\end{figure}

\paragraph{Interpretation.}  
While these normalization strategies enable comparability across heterogeneous tasks, they also introduce some limitations.  
For min-max scaling, small score differences may be exaggerated, especially in datasets with few languages.  
For $z$-score normalization, languages that appear only in relatively harder datasets may be penalized more strongly, making this approach less suitable for aggregation across tasks. Accordingly, we interpret both the aggregated scores and the heatmaps as indicators of overall trends rather than exact performance values.  We note that individual outliers may not fully reflect a language’s true capabilities, particularly when it appears in only a limited number of datasets. Figure~\ref{fig:combined_heatmap} illustrates these normalization strategies, 
providing a compact graphical overview of our results. Further sections continue to the detailed, dataset-specific analyses.

\subsubsection{MultiEURLEX}
\begin{figure}[H]
    \centering
    \includegraphics[width=0.9\linewidth]{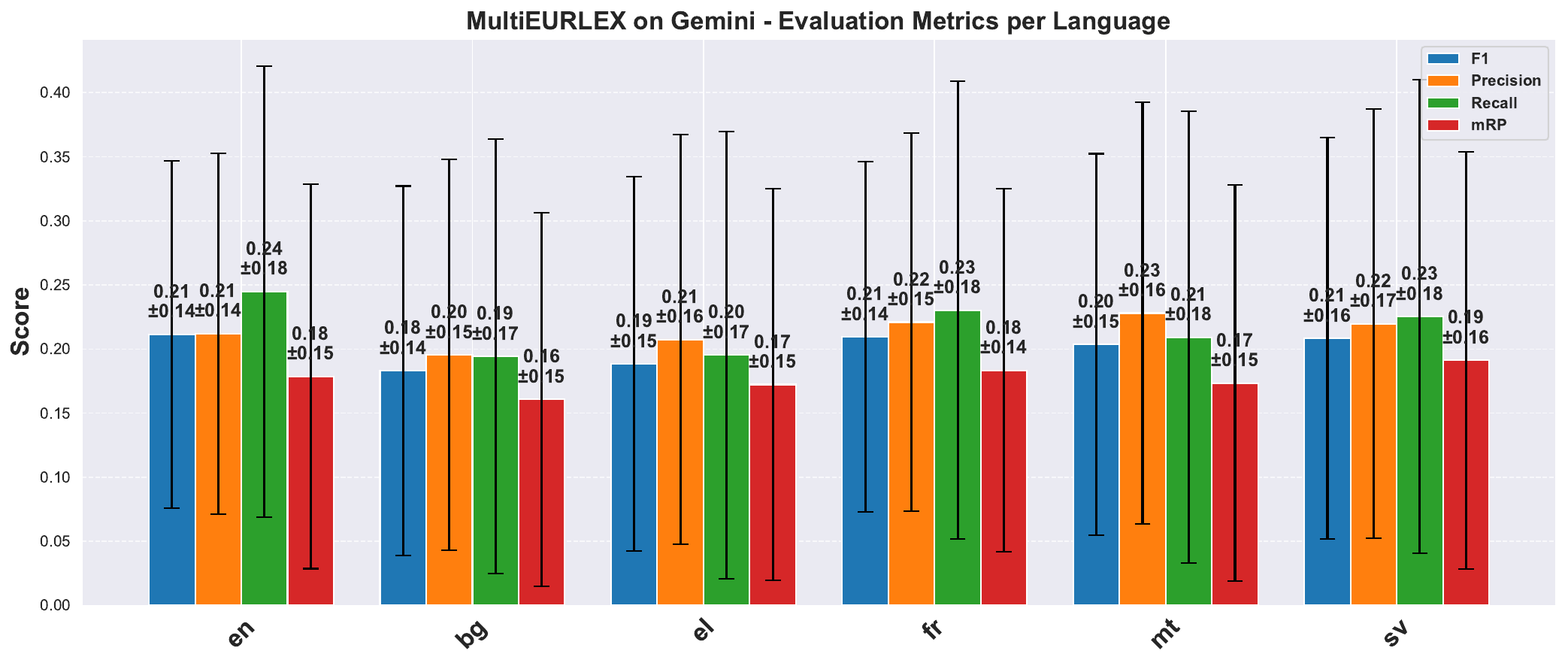}
    \caption{Evaluation metrics per language on the MultiEURLEX dataset with Gemini 1.5 Flash. Error bars show standard deviation across examples.}
    \label{fig:multi_eurlex_metrics}
\end{figure}

For the evaluation of the MultiEURLEX dataset, we use the \textbf{Gemini 1.5 Flash} model. The experiment is conducted on 200 data points for each of six representative languages: English, Bulgarian, Greek, French, Maltese, and Swedish. While the dataset spans 23 languages, we restrict our analysis to this subset in order to capture variation across different language families, alphabets, and resource availability, while keeping the evaluation computationally feasible.  

Figure~\ref{fig:multi_eurlex_metrics} presents the results, reporting Precision, Recall, F1 score, and Mean-r Precision. While most languages yield comparable results, the high variance observed renders the findings less reliable. This can be attributed to the inherent difficulty of the task posed by the MultiEURLEX dataset: a multi-label classification problem involving 567 labels. Such complexity poses a significant challenge for commercial LLMs. These results suggest that commercial LLMs, when used out-of-the-box and without task-specific pretraining, struggle with domain-specific tasks, such as legal classification, especially when the task is highly complex, regardless of the language.

\subsubsection{Eur-Lex-Sum}
For the evaluation of the Eur-Lex-Sum dataset, we use the \textbf{LLaMA 3.1–8B} model.

\begin{figure}[H]
    \centering
    \includegraphics[width=0.9\linewidth]{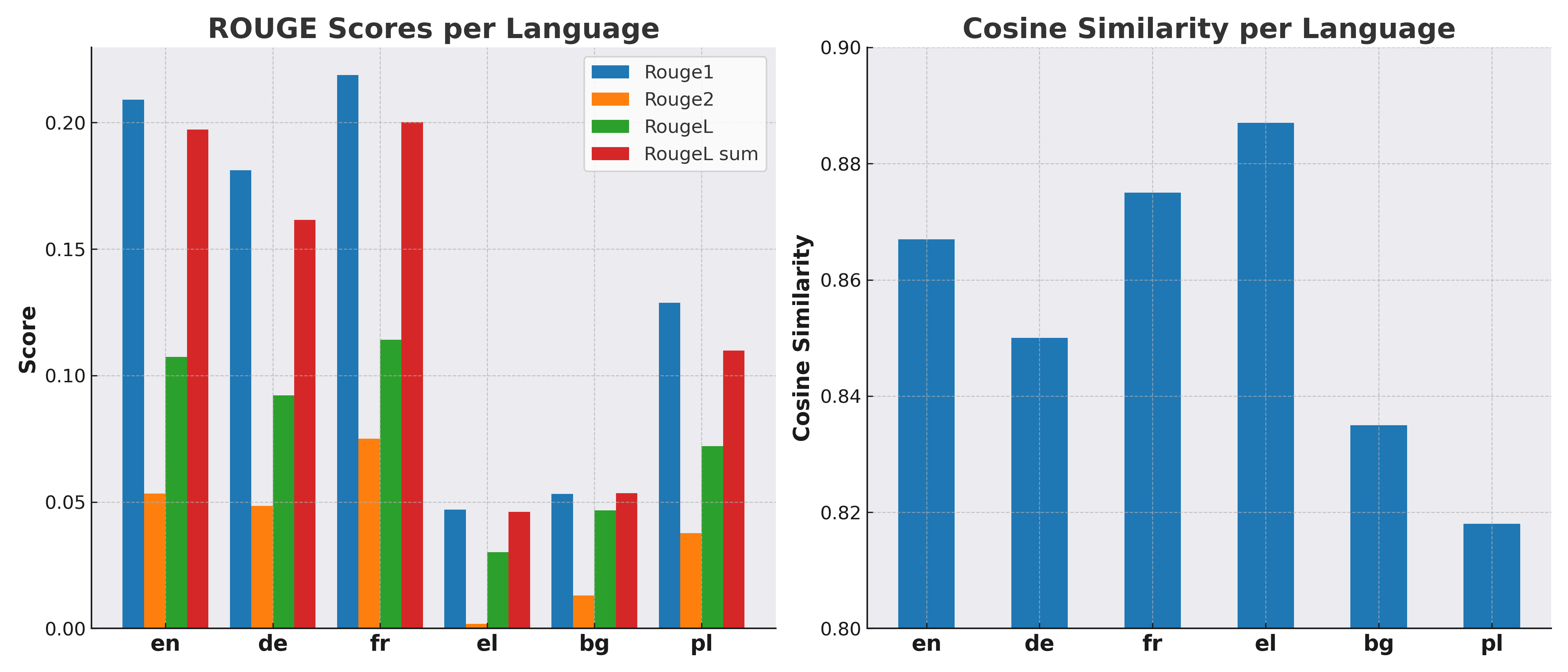}
    \caption{Rouge scores per language on the left and cosine similarities on the right. These results are from 25 items per language on the Eur-Lex-Sum dataset.}
    \label{fig:eurlexsum-llama}
\end{figure}

To evaluate LLaMA on the Eur-Lex-Sum dataset we initially calculate ROUGE scores of its summaries, as shown in Figure~\ref{fig:eurlexsum-llama}. While the scores appear relatively low, this reflects the fact that ROUGE is a strict metric relying on exact lexical overlap rather than contextualized analysis. To provide a complementary perspective, we also compute cosine similarities using BERT embeddings~\cite{devlin2019bertpretrainingdeepbidirectional}, which better capture semantic relatedness.

The sub-figures in Figure~\ref{fig:eurlexsum-llama} show that the model performs best in French under ROUGE, with English following closely behind and German slightly lower. By contrast, under cosine similarity Greek moves from being the lowest under ROUGE to the highest, while Polish drops to the bottom. This highlights the difference between lexical and semantic evaluation: Greek summaries capture meaning despite low lexical overlap, whereas Polish shows the opposite tendency. Nevertheless, in legal tasks precise terminology is crucial, and we therefore place greater weight on ROUGE-L Sum as the more appropriate metric in this domain.

\subsubsection{EUROPA}
\begin{figure}[H]
    \centering
    \includegraphics[width=0.45\linewidth]{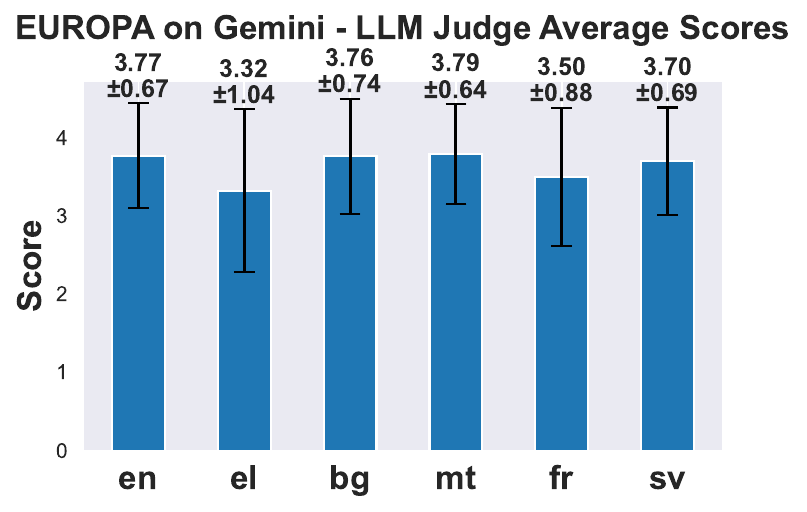}
    \caption{Mean LLM Score (on a 1–5 scale) per language on the EUROPA dataset with Gemini 1.5 Flash. Error bars show standard deviation across examples.}
    \label{fig:europa_score}
\end{figure}

For the evaluation of the EUROPA dataset, we use the \textbf{Gemini 1.5 Flash} model.  The experiment is conducted on 200 data points for each of six representative languages: English, Greek, Bulgarian, Maltese, French, and Swedish. Although the dataset covers 24 languages, we limit our evaluation to this subset in order to balance computational cost with linguistic diversity, ensuring coverage across different families, alphabets, and levels of resource availability. 

The results are presented in Figure~\ref{fig:europa_score}, which reports the mean LLM Judge Scores for each language. English (en) demonstrates performance comparable to Bulgarian (bg) and Maltese (mt), followed by slightly lower scores for Swedish (sv), French (fr), and finally Greek (el). While the task, generating key phrases for a given text, is not particularly complex, it remains situated within the legal domain. The results suggest that Gemini performs relatively consistently across languages in nuanced, domain-specific contexts such as law. This may be attributed to the model’s ability to generalize semantic understanding across languages, particularly when the task involves short and structured outputs like key phrases. However, its performance is still constrained by the legal nature of the texts, which pose domain-specific challenges that limit the model’s effectiveness.

\subsubsection{Online Terms of Service}
\label{tos_section}
For the evaluation of the Online Terms of Service (ToS) dataset, we use the \textbf{Gemini 1.5 Flash} model. The experiment is conducted using 300 data points for each language and three different prompts, each more assertive than the previous one.

\paragraph{Basic Prompt}\mbox{}\\

\begin{figure}[H]
    \centering
    \includegraphics[width=0.5\linewidth]{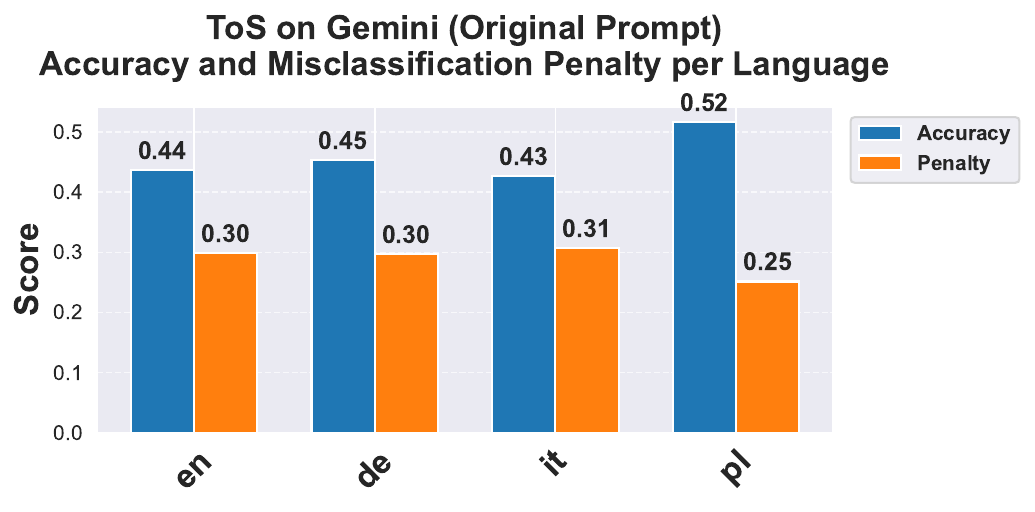}
    \caption{Average accuracy and penalty score (both on a 0–1 scale) per language on the Online Terms of Service dataset, with Gemini 1.5 Flash, using the basic prompt. The penalty score is calculated as described in Section~\ref{datasets}.}
    \label{fig:tos_scores_original}
\end{figure}

Initially, a basic instruction is provided to the model:

\begin{quote}
\texttt{You are a legal document fairness classifier. Above is a clause from an online Terms of Service document.\\
Your task is to classify the fairness of the clause based on its content.\\
- Only select one of the following labels:\\
0: clearly fair\\
1: potentially unfair\\
2: clearly unfair\\
Return \textbf{only the label number}.\\
- Do not explain your answer or include any other text.}
\end{quote}

Figure~\ref{fig:tos_scores_original} presents the resulting accuracy and penalty scores. English (en) exhibits performance comparable to German (de) and Italian (it), with Polish (pl) performing slightly better. Nevertheless, overall accuracy remains relatively low for what is ostensibly a simple three-label classification task.

However, an analysis of the confusion matrices in Figure~\ref{fig:confusion_matrices_original} (Appendix~\ref{appendix:graphs}) reveals a consistent pattern: many misclassifications involve underestimating unfairness. For example, instances labeled as \textit{potentially unfair} are frequently predicted as \textit{clearly fair}, and those labeled as \textit{clearly unfair} are often predicted as \textit{potentially unfair}. A likely explanation is that general-purpose LLMs, lacking legal grounding, struggle with fine-grained distinctions in fairness-related judgments and tend to default to vague or ``safe'' responses. Prior research~\cite{scherrerEvaluatingMoralBeliefs2023} supports this interpretation, showing that LLMs often express uncertainty when faced with ambiguous scenarios. Furthermore, models trained predominantly on public web content may exhibit a tendency to avoid accusatory or critical language~\cite{salechaLargeLanguageModels2024}, resulting in an overprediction of the \textit{clearly fair} label and an underprediction of \textit{clearly unfair} cases.

\paragraph{More Assertive Prompt}\mbox{}\\

\begin{figure}[H]
    \centering
    \includegraphics[width=0.5\linewidth]{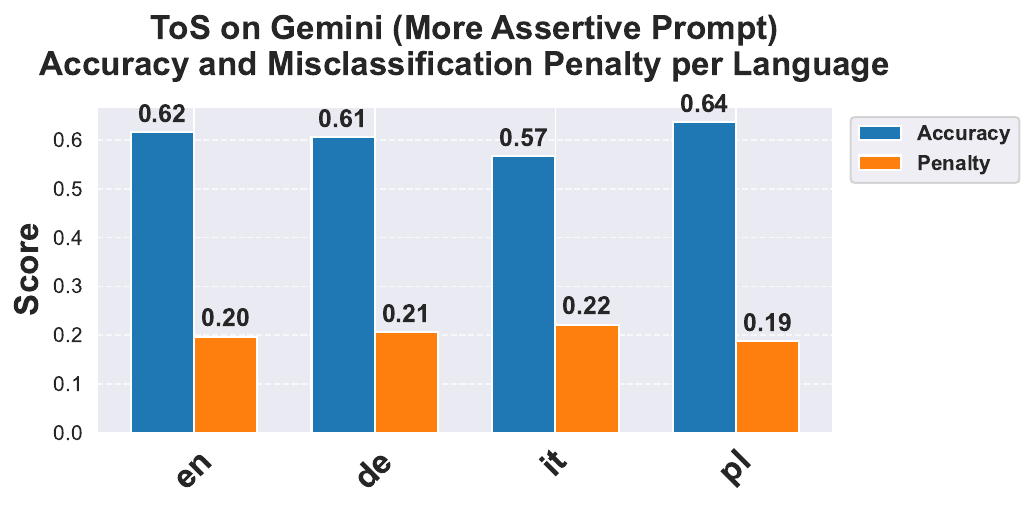}
    \caption{Average accuracy and penalty score (both on a 0–1 scale) per language on the Online Terms of Service dataset, with Gemini 1.5 Flash, using the more assertive prompt. The penalty score is calculated as described in Section~\ref{datasets}.}
    \label{fig:tos_scores_assertive}
\end{figure}

To counteract the hesitancy observed, we adjust the prompt to be more assertive, encouraging the model to focus on legal implications and disregard tone or phrasing:

\begin{quote}
\texttt{You are a legal document fairness classifier. Above is a clause from an online Terms of Service document.\\
Your task is to classify the fairness of the clause \textcolor{ForestGreen}{based strictly on its legal implications and potential consumer impact.}\\
\textcolor{ForestGreen}{Do not let tone, phrasing, or politeness influence your decision.}\\
\textcolor{ForestGreen}{Be objective and impartial — base your judgment only on how the clause affects users' rights or obligations.}\\
- Only select one of the following labels:\\
0: clearly fair\\
1: potentially unfair\\
2: clearly unfair\\
Return \textbf{only the label number}.\\
- Do not explain your answer or include any other text.}
\end{quote}

As shown in Figure~\ref{fig:tos_scores_assertive}, this adjustment leads to more confident classifications and improved performance, with all languages exhibiting significantly higher accuracy and lower penalty scores. However, the confusion matrices in Figure~\ref{fig:confusion_matrices_assertive} (Appendix~\ref{appendix:graphs}) reveal that predictions of \textit{clearly unfair} remain relatively rare, indicating a persistent bias toward moderate classifications.

\paragraph{Highly Assertive Prompt}\mbox{}\\

\begin{figure}[H]
    \centering
    \includegraphics[width=0.5\linewidth]{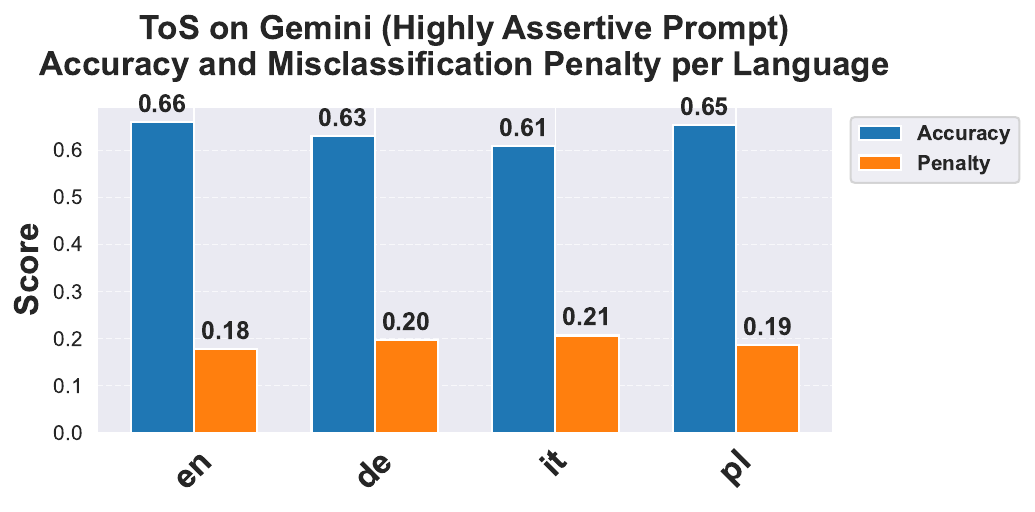}
    \caption{Average accuracy and penalty score (both on a 0–1 scale) per language on the Online Terms of Service dataset, with Gemini 1.5 Flash, using the highly assertive prompt. The penalty score is calculated as described in Section~\ref{datasets}.}
    \label{fig:tos_scores_highly_assertive}
\end{figure}

Finally, we craft an even more assertive prompt to explicitly encourage decisive judgments in cases of imbalance:

\begin{quote}
\texttt{You are a legal document fairness classifier. Above is a clause from an online Terms of Service document.\\
Your task is to classify the fairness of the clause \textcolor{ForestGreen}{strictly based on its legal implications and potential consumer impact.}\\
\textcolor{ForestGreen}{- Do not let tone, phrasing, or politeness influence your decision.}\\
\textcolor{ForestGreen}{- Be objective and impartial — focus solely on how the clause affects users’ rights and obligations.}\\
\textcolor{ForestGreen}{- Do not hesitate to select a strong classification if the clause imposes a significant imbalance or limitation.}\\
- Only select one of the following labels:\\
0: clearly fair\\
1: potentially unfair\\
2: clearly unfair\\
Return \textbf{only the label number}.\\
- Do not explain your answer or include any other text.}
\end{quote}

As illustrated in Figure~\ref{fig:tos_scores_highly_assertive}, this modification results in a slight increase in accuracy and a decrease in penalty scores across all languages. English (en) now shows performance comparable to the previously dominant Polish (pl), while German (de) and Italian (it) nearly reach the same level. Notably, English exhibits the largest improvement, with a 50\% increase in accuracy and a 40\% reduction in penalty score between the least and most assertive prompts. German achieves a 40\% increase in accuracy and a 33\% decrease in penalty score; Italian sees a 42\% increase and 32\% decrease, while Polish improves by 25\% and 24\%, respectively. These results suggest that English is the most sensitive to prompt modifications and responds most positively to increased assertiveness.

The confusion matrices in Figure~\ref{fig:confusion_matrices_highly_assertive} (Appendix~\ref{appendix:graphs}) indicate more balanced outcomes under the highly assertive prompt. While some \textit{clearly unfair} cases continue to be predicted as \textit{potentially unfair}, there is also notable shifts in the other direction. For example, some instances previously incorrectly labeled as \textit{potentially unfair} are now correctly classified as \textit{clearly unfair} and some that had been incorrectly labeled as \textit{clearly fair} are correctly reassigned to \textit{potentially unfair}. Additionally, some instances previously correctly classified as \textit{potentially unfair} are now predicted as \textit{clearly unfair}. Although this more assertive prompting occasionally leads the model to overcorrect, misclassifying previously correct instances by predicting a more negative label, it largely mitigates the earlier issue of rarely predicting the \textit{clearly unfair} category. Overall, the model demonstrates greater decisiveness and a more balanced distribution across all fairness labels.

These findings indicate that LLMs, particularly in English, may exhibit a bias toward safer or more positive classifications. They also highlight the importance of prompt engineering: well-structured and specific prompts can significantly affect model behavior and improve performance in sensitive, domain-specific tasks.

\subsubsection{LEXam}
\begin{figure}[H]
    \centering
    \includegraphics[width=0.45\linewidth]{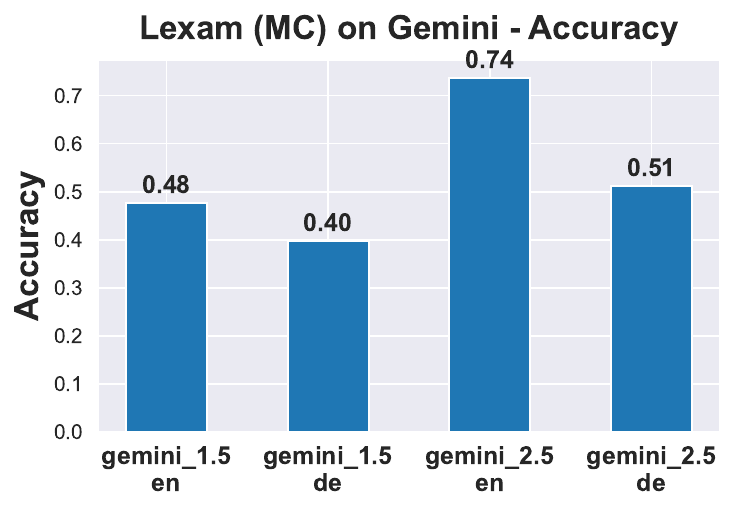}
    \caption{Mean accuracy per language on the LEXam dataset (on the multiple choice task) with Gemini 1.5 Flash and Gemini 2.5 Flash. }
    \label{fig:lexam_mc}
\end{figure}
\begin{figure}[H]
    \centering
    \includegraphics[width=0.45\linewidth]{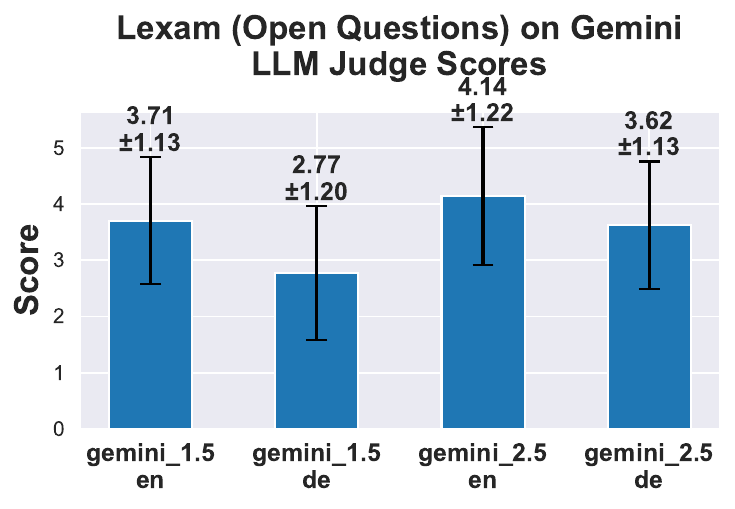}
    \caption{Mean LLM Score (on a 1–5 scale) per language on the LEXam dataset (on the open questions task) with Gemini 1.5 Flash and Gemini 2.5 Flash. Error bars show standard deviation across examples.}
    \label{fig:lexam_open}
\end{figure}

For the evaluation of the LEXam dataset, we use the  \textbf{Gemini 1.5 Flash} and \textbf{Gemini 2.5 Flash (Preview 05/20)} models. We select both models to compare the performance of a widely adopted production-grade LLM (1.5 Flash) with its next-generation successor (2.5 Flash Preview), enabling an assessment of progress in multilingual and reasoning context-specific capabilities. For the multiple-choice task, the experiment is conducted using 500 data points per language, while for the open-ended questions task, 400 data points are used per language.

Figure \ref{fig:lexam_mc} illustrates the performance of both models on the 4-choice questions. Figure \ref{fig:lexam_open} presents the mean LLM Judge score per language for each model in the open-ended task. In both tasks, the models perform better on the English (en) version than on the German (de) one.

Both figures demonstrate a clear performance improvement from Gemini 1.5 Flash to Gemini 2.5 Flash. Notably, the German performance on Gemini 2.5 is comparable to the English performance on Gemini 1.5, while English performance on 2.5 significantly surpasses that of 1.5, especially in the multiple-choice task.

Despite the substantial performance improvements of Gemini 2.5 Flash over its predecessor, we observe some undesirable behaviors during evaluation. Specifically, for the multiple-choice task, we require the model to return answers in a strict predefined format. While Gemini 1.5 Flash consistently adheres to this format, Gemini 2.5 Flash frequently deviates, although it still provides human-readable responses. Consequently, some manual post-processing is necessary to extract the selected options.

Furthermore, in the open-ended English questions, Gemini 2.5 Flash occasionally responds in German, despite being explicitly instructed to answer in English. This behavior likely stems from the presence of German legal terminology in the tasks, given the Swiss legal exam context of the dataset. These issues are not observed with Gemini 1.5 Flash.

Gemini demonstrates significantly stronger performance on English tasks compared to multilingual ones. Moreover, the results indicate a clear improvement across model generations, particularly in specialized domains such as law, for both English and multilingual settings. Nevertheless, the inability of Gemini 2.5 Flash to consistently follow explicit formatting instructions raises concerns about its reliability in tasks requiring strict adherence to prompts.

\subsubsection{XNLI}
For the evaluation of the XNLI dataset, we use the \textbf{Gemini 1.5 Flash} and \textbf{LLaMA 3.2–3B} models.

\paragraph{Evaluation with Gemini 1.5 Flash}\mbox{}\\
\begin{figure}[H]
    \centering
    \includegraphics[width=0.45\linewidth]{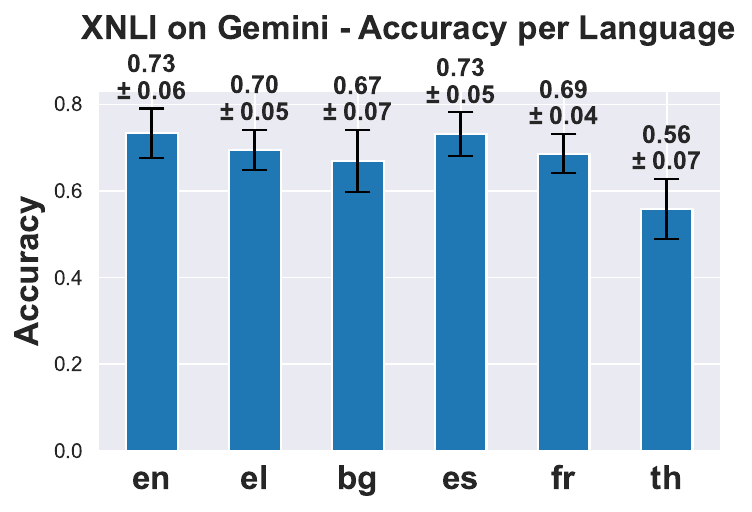}
    \caption{Mean accuracy per language on the XNLI dataset with Gemini 1.5 Flash. Error bars show standard deviation computed from per-example correctness (1 = correct, 0 = incorrect) across 25 runs, reflecting variability in predictions across runs.}
    \label{fig:xnli_accuracy}
\end{figure}

\begin{figure}[H]
    \centering
    \includegraphics[width=0.9\linewidth]{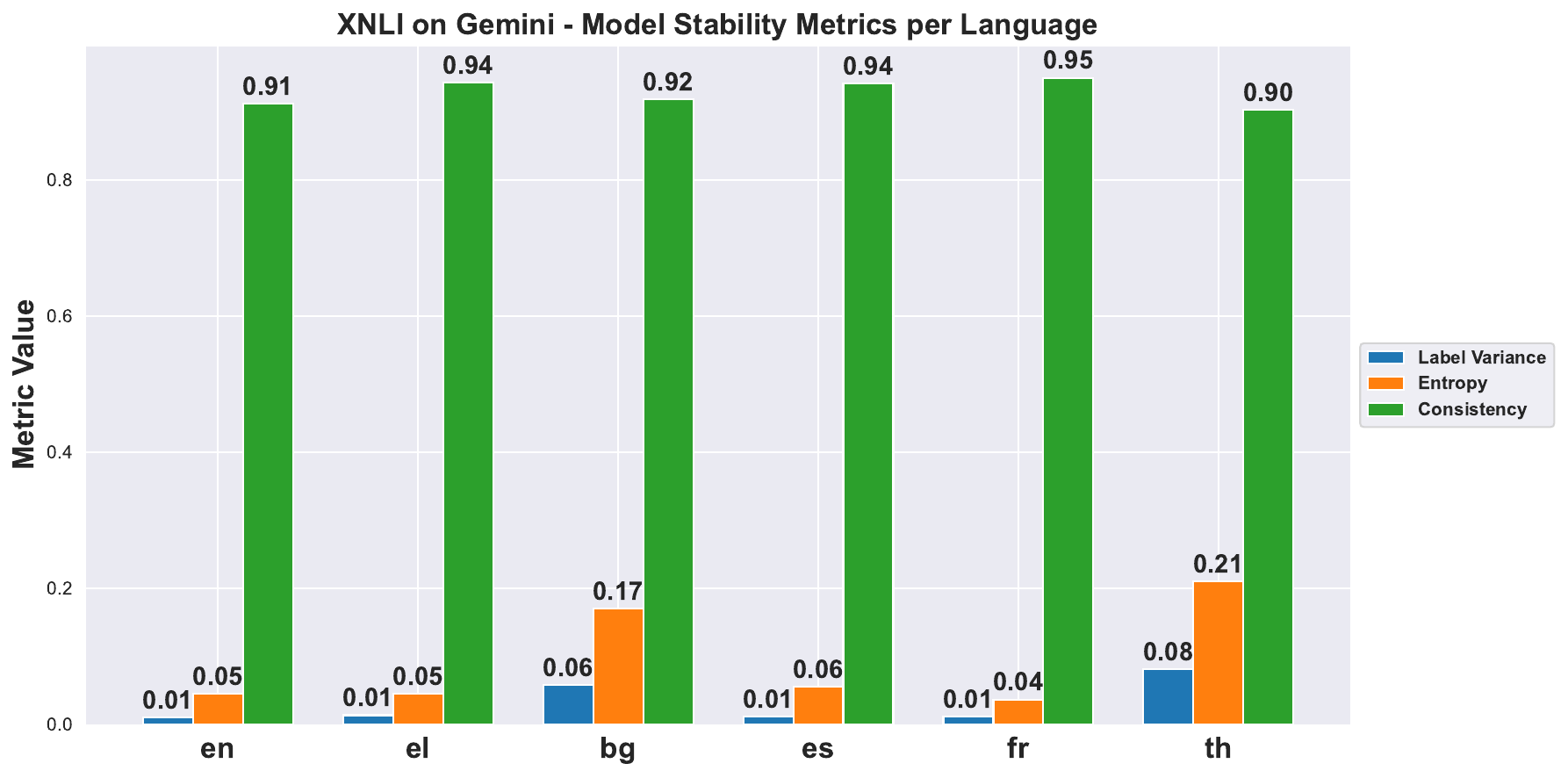}
    \caption{
    Stability metrics per language on the XNLI dataset with Gemini 1.5 Flash. 
    Label variance reflects variability in predicted class labels across runs. Entropy measures prediction uncertainty across labels. Consistency indicates the proportion of predictions that agree with the majority vote. 
    These metrics together characterize the robustness and stability of model predictions.
    }
    \label{fig:xnli_stability}
\end{figure}

The experiment with \textbf{Gemini 1.5 Flash} is conducted over 25 independent runs, each using the same 200 data points per language.

Figure~\ref{fig:xnli_accuracy} reports the mean accuracy per language, along with the standard deviation, which captures variability in per-example correctness (1 = correct, 0 = incorrect) across runs. Once again, we observe comparable performance across most languages: English (en) and Spanish (es) lead with an accuracy of 73\%, followed by Greek (el) at 70\%, French (fr) at 69\%, Bulgarian (bg) at 67\%, and, with a notable drop, Thai (th) at 56\%.

Figure~\ref{fig:xnli_stability} presents stability metrics, including the variance in predicted labels (or correctness) across runs for each individual input. This offers a more fine-grained view of prediction consistency beyond aggregate accuracy. All languages, with the exception of Bulgarian and Thai, exhibit similar stability. In contrast, Bulgarian and Thai show significantly higher label variance and entropy.

These results demonstrate Gemini’s ability to perform both accurately and consistently across most languages in general-purpose, non-specialized tasks. However, the notably lower accuracy and higher prediction variance observed for Thai suggest that the model may be less robust when handling languages that are underrepresented in its training data or that differ significantly in structure from more widely spoken Indo-European languages.

\paragraph{Evaluation with LLaMA 3.2–3B}\mbox{}\\
\begin{figure}[H]
    \centering
    \includegraphics[width=0.5\linewidth]{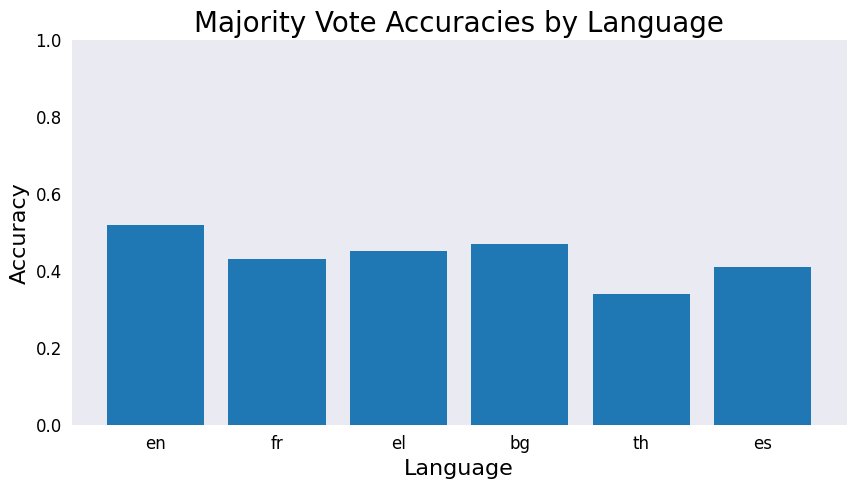}
    \caption{
    Accuracy of LLaMA per language. The labels used were the majority votes from 30 independent runs and 300 data points per run.
    }
    \label{fig:llama_accuracy_xnli}
\end{figure}

\begin{figure}[H]
    \centering
    \includegraphics[width=0.9\linewidth]{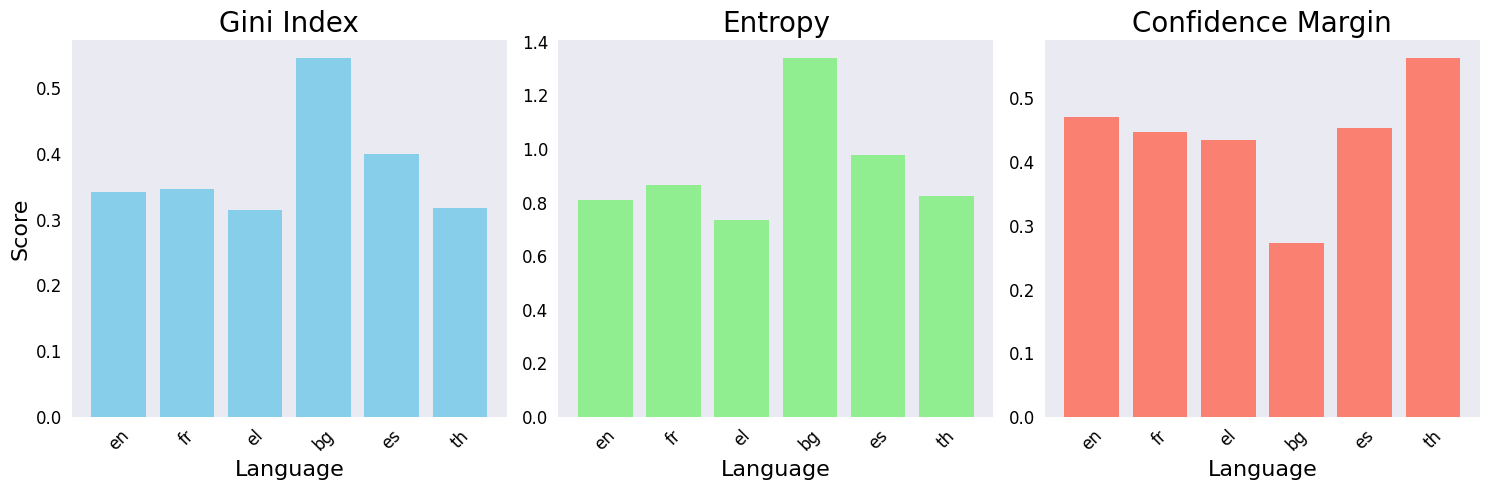}
    \caption{
    Uncertainty measures of LLaMA per language. These results are based on 30 independent runs and 300 data points per run.
    }
    \label{fig:llama_uncertainty_xnli}
\end{figure}

\begin{figure}[H]
    \centering
    \includegraphics[width=0.9\linewidth]{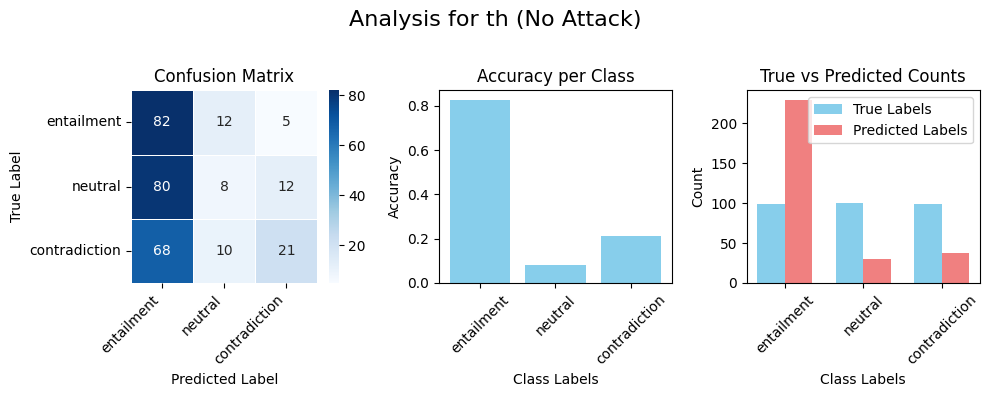}
    \caption{
    Analysis for the performance of LLaMA in Thai. These results are based on 30 independent runs and 300 data points per run.
    }
    \label{fig:llama_th}
\end{figure}

For the evaluation of LLaMA we conduct 30 independent runs on 300 data points of the XNLI dataset. Figure \ref{fig:llama_accuracy_xnli} shows the accuracy of LLaMA on each language. The predicted label is determined by majority vote. It is evident that, although marginally, LLaMA performs better in English than all other evaluated languages. It is also important to note, that for Thai, which is the least similar and likely least trained on language, the model performs significantly worse than all other languages. 

Despite the findings based on the accuracy metric, it is also crucial to understand the stability of the model and its variability. Figure \ref{fig:llama_uncertainty_xnli} shows uncertainty measures on the same runs. We use gini index and entropy to measure the impurity of each cluster of responses by the model and we use the confidence margin, which is calculated as the difference between the most commonly and second most commonly guessed labels, to further understand how consistent the model is with its top answer. From the graphs we can see something quite surprising, that is better consistency in Thai than English. This does not align with our initial expectations of higher stability in higher resource languages. We see that Thai is 2nd in gini index, 3rd in entropy and first, by quite a significant margin, in the confidence margin metric. It is important to mention, however, that the model is more consistently wrong in Thai as its accuracy is last among all languages. 
\par In order to understand further the results and why Thai seems to be more stable and consistent we look into the analysis for Thai alone as presented in figure \ref{fig:llama_th}. From first glance it becomes very clear that the model does almost no better than just guessing entailment always, which is also seen in the accuracy since it is close to random(i.e. 33\%). The accuracies per class plot shows very high accuracy in entailment, as it is extremely overguessed and very low accuracies for the other two classes. It is difficult to really understand why the model behaves this way but our best estimation is that when unsure the model guesses entailment as a safe option. 

\paragraph{Cross-model comparison}\mbox{}\\

When comparing Gemini 1.5 Flash and LLaMA 3.2–3B on XNLI, we observe both expected and surprising behaviors. For both models, English emerges as the strongest language in terms of accuracy (tied with Spanish in Gemini), while Thai consistently ranks lowest, most probably reflecting its typological distance from English and relative underrepresentation in training data.

In terms of absolute performance, Gemini consistently outperforms LLaMA by a wide margin: the gap ranges from 20 points for Bulgarian (67\% vs. 47\%) to as high as 32 points for Spanish (73\% vs. 41\%). On average across all overlapping languages, Gemini scores about 24 percentage points higher than LLaMA, demonstrating its much stronger capabilities both in English and multilingual settings.

The largest divergence between the two models' performance occurs for Spanish: while Gemini places it at the very top alongside English (73\%), LLaMA ranks it second-to-last with only 41\% accuracy. This is unexpected, because Spanish is a high-resource language with relatively large amount of training data and closely related to English. This result highlights how differently models can leverage cross-lingual transfer and could suggest that Gemini exhibits substantially stronger cross-lingual transferability than LLaMA.

Generally, these findings align with our expectations that Gemini which is a larger proprietary model trained on broader multilingual data exhibits greater performance across all languages. At the same time, the Spanish anomaly in LLaMA highlights that model behavior is not always predictable from language relatedness or resource availability, underlining the need of empirical evaluation across multiple models.

\subsubsection{XQuAD}
\begin{figure}[H]
    \centering
    \includegraphics[width=0.6\linewidth]{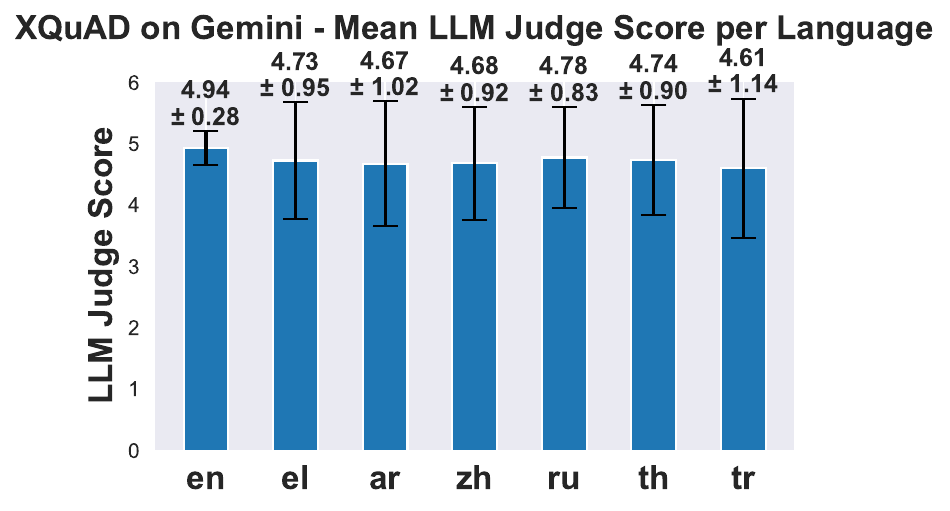}
    \caption{Mean LLM Score (on a 1–5 scale) per language on the XQuAD dataset with Gemini 1.5 Flash. Error bars show standard deviation across examples.}
    \label{fig:xquad_score}
\end{figure}

For the evaluation of the XQuAD dataset, we use the \textbf{Gemini 1.5 Flash} model. The experiment is
conducted using 200 data points for each language. 

Figure~\ref{fig:xquad_score} reports the mean LLM Judge Scores across languages. Additional metrics (BLEU, METEOR, and cosine similarity) are presented in Figure~\ref{fig:xquad_metrics} in Appendix~\ref{appendix:graphs}, but due to their high variance, we select the LLM Judge Score as the primary evaluation metric for this task. As shown in Figure~\ref{fig:xquad_score}, all languages perform well, which is expected given that this is the simplest task evaluated in this paper, requiring answers to straightforward questions based on a short given passage.

English achieves the highest score, with an almost perfect mean of 5 and substantially lower variance, indicating strong consistency across individual examples. Following English are Russian (ru), Thai (th), and Greek (el), then Arabic (ar), and finally Turkish (tr), with a performance gap of 0.17 between the highest-performing non-English language (Russian) and the lowest (Turkish).

These results suggest that Gemini performs very well on simple, non-domain-specific tasks that do not require complex reasoning, even in a multilingual setting. Notably, the near-perfect performance on English highlights the model’s strong capabilities when operating in its most likely training-dominant language on simple tasks.

\subsection{Adversarial Robustness Evaluation}

\subsubsection{LEXam}
\begin{figure}[H]
    \centering
    \includegraphics[width=0.55\linewidth]{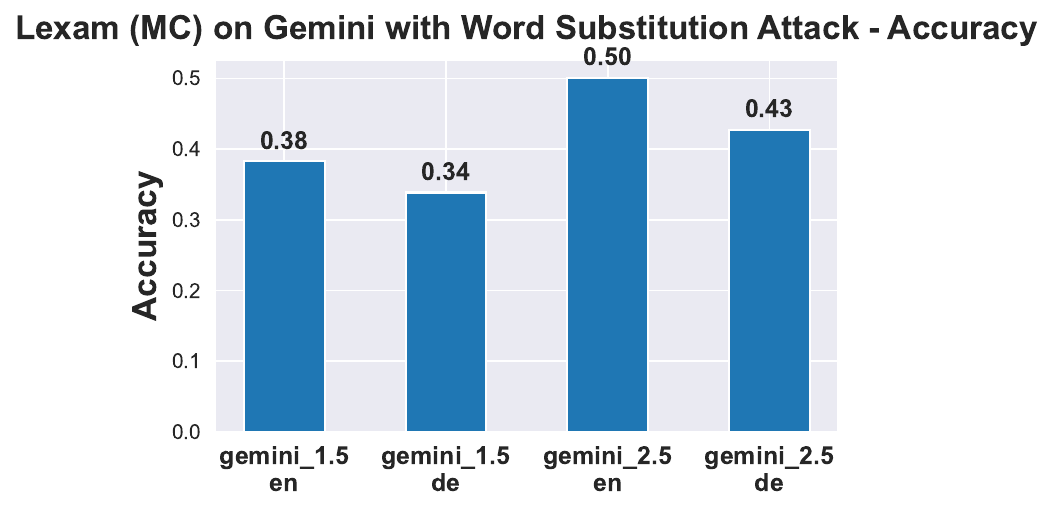}
    \caption{Mean accuracy per language on the LEXam dataset (on the multiple choice task) with Gemini 1.5 Flash and Gemini 2.5 Flash, after applying a BERT-based word substitution attack. }
    \label{fig:lexam_mc_attack}
\end{figure}
\begin{figure}[H]
    \centering
    \includegraphics[width=0.6\linewidth]{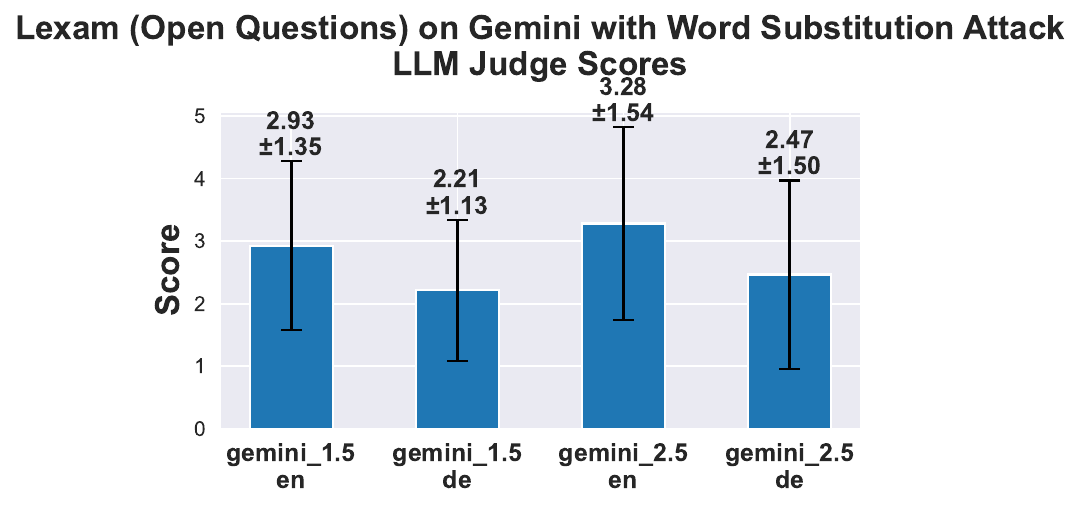}
    \caption{Mean LLM Score (on a 1–5 scale) per language on the LEXam dataset (on the open questions task) with Gemini 1.5 Flash and Gemini 2.5 Flash, after applying a BERT-based word substitution attack. Error bars show standard deviation across examples.}
    \label{fig:lexam_open_attack}
\end{figure}

To evaluate the adversarial robustness of \textit{Gemini 1.5 Flash} and \textit{Gemini 2.5 Flash} on LEXam, we apply a BERT-based contextual word substitution attack, implemented using the \texttt{nlpaug} library, on the same data points previously evaluated under standard conditions. Specifically, we use the \texttt{ContextualWordEmbsAug} augmenter with language-specific Bert models, operating in substitution mode. Each input is perturbed with a probability of 15\% per token (\texttt{aug\_p = 0.15}), replacing words with semantically similar alternatives based on contextual embeddings.

Figure \ref{fig:lexam_mc_attack} illustrates the performance of both models on the 4-choice questions, while Figure \ref{fig:lexam_open_attack} shows the mean LLM Judge score per language for the open-ended task. As expected, both tasks exhibit a decline in performance compared to the non-attacked setting, while the trends stay similar to the non-attacked setting.

The adversarial attack significantly reduces model performance, particularly on the open-ended questions, where the variance increases for most cases, suggesting higher output instability under perturbation. Nevertheless, Gemini 2.5 Flash continues to perform relatively well, especially on the multiple-choice task, given the adversarial setting. Additionally, the open-ended English responses for both models maintain a relatively strong performance despite the adversarial conditions.

Since the results show improvement on \textit{Gemini~2.5~Flash}, we also employ a random character insertion attack on it. Specifically, we use the \texttt{RandomCharAug} augmenter from the \texttt{nlpaug} library, which performs character-level insertions with a probability of 30\% per character (\texttt{aug\_char\_p = 0.3}). What this means is that for each character position in the input text, there is a 30\% chance that a random character will be inserted immediately before or after that position. This introduces noise by randomly inserting characters into the input text, simulating low-level adversarial perturbations.

\begin{figure}[H]
    \centering
    \includegraphics[width=0.55\linewidth]{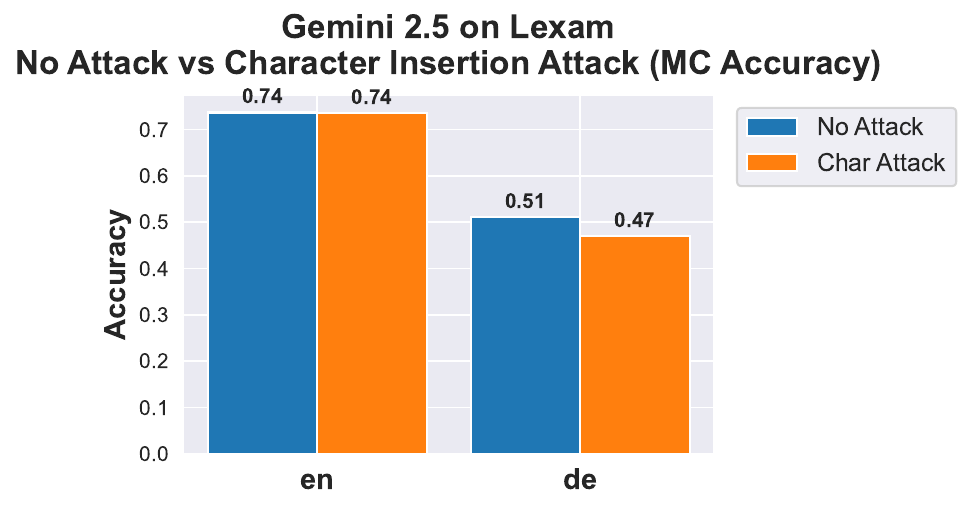}
    \caption{Mean accuracy per language on the LEXam dataset (on the multiple choice task) with Gemini 2.5 Flash, after applying a random character insertion attack.}
    \label{fig:lexam_mc_char_attack}
\end{figure}
\begin{figure}[H]
    \centering
    \includegraphics[width=0.6\linewidth]{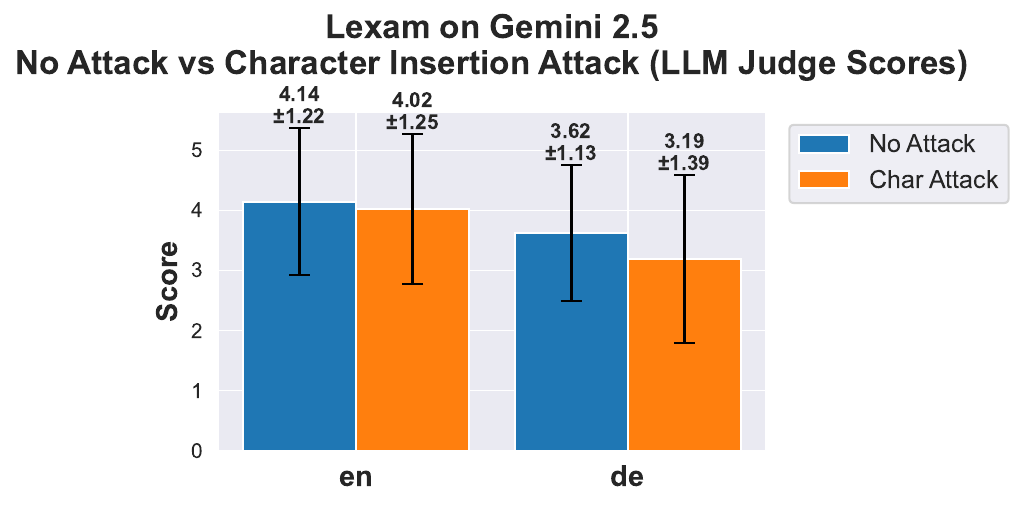}
    \caption{Mean LLM Score (on a 1–5 scale) per language on the LEXam dataset (on the open questions task) with Gemini 2.5 Flash, after applying a random character insertion attack. Error bars show standard deviation across examples.}
    \label{fig:lexam_open_char_attack}
\end{figure}

Figure~\ref{fig:lexam_mc_char_attack} shows the results of the character-level attack on the multiple-choice task, while Figure~\ref{fig:lexam_open_char_attack} presents the corresponding results on the open-ended task. For English, the performance on the multiple-choice task remains unchanged, while on the open-ended task, the score drops by a very small amount (0.12 points), and the standard deviation decreases slightly by 0.03.

In German, although the impact is not drastic, the performance degradation is more noticeable than in English. Specifically, there is a 0.04 drop in accuracy on the multiple-choice task and a 0.43-point reduction in the open-ended task score, accompanied by a 0.26 increase in standard deviation.

Overall, these results highlight two key findings. First, Gemini 2.5 Flash demonstrates a notable degree of robustness under adversarial input, particularly on the multiple-choice task. Second both the open-ended question results and the outcomes of the character-level attack indicate that English-language inputs consistently yield stronger performance, even under attack, reinforcing the models' relative strength in English compared to other languages.

\subsubsection{XNLI}

\paragraph{Evaluation with Gemini 1.5 Flash}\mbox{}\\

\begin{figure}[H]
    \centering
    \includegraphics[width=0.6\linewidth]{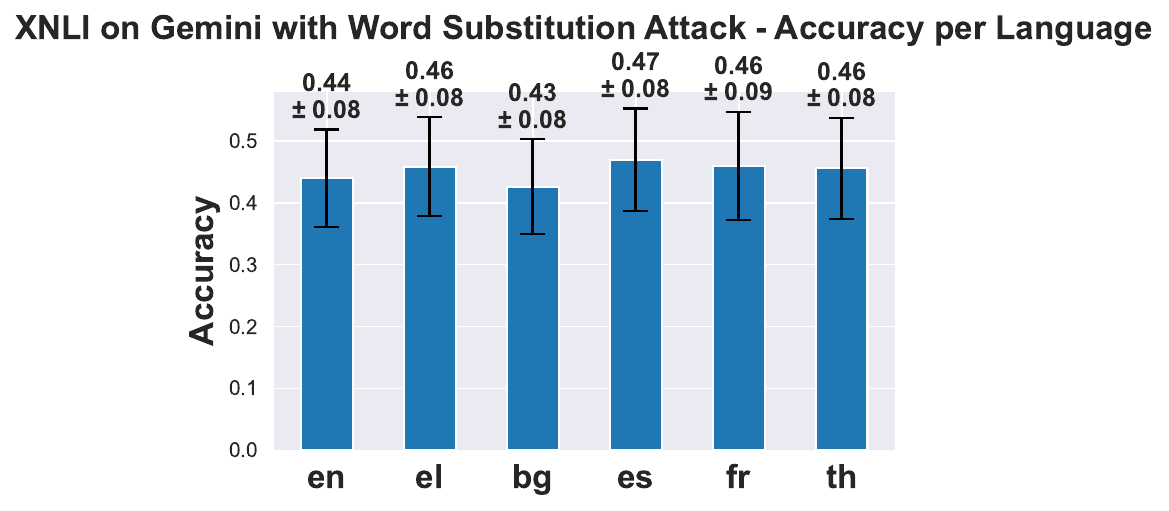}
    \caption{
    Mean accuracy per language on the XNLI dataset with Gemini 1.5 Flash, after applying a BERT-based word substitution attack. 
    Error bars show standard deviation computed from per-example correctness (1 = correct, 0 = incorrect) across 25 runs, reflecting variability in predictions across runs.
    }
    \label{fig:xnli_accuracy_attack}
\end{figure}

\begin{figure}[H]
    \centering
    \includegraphics[width=0.9\linewidth]{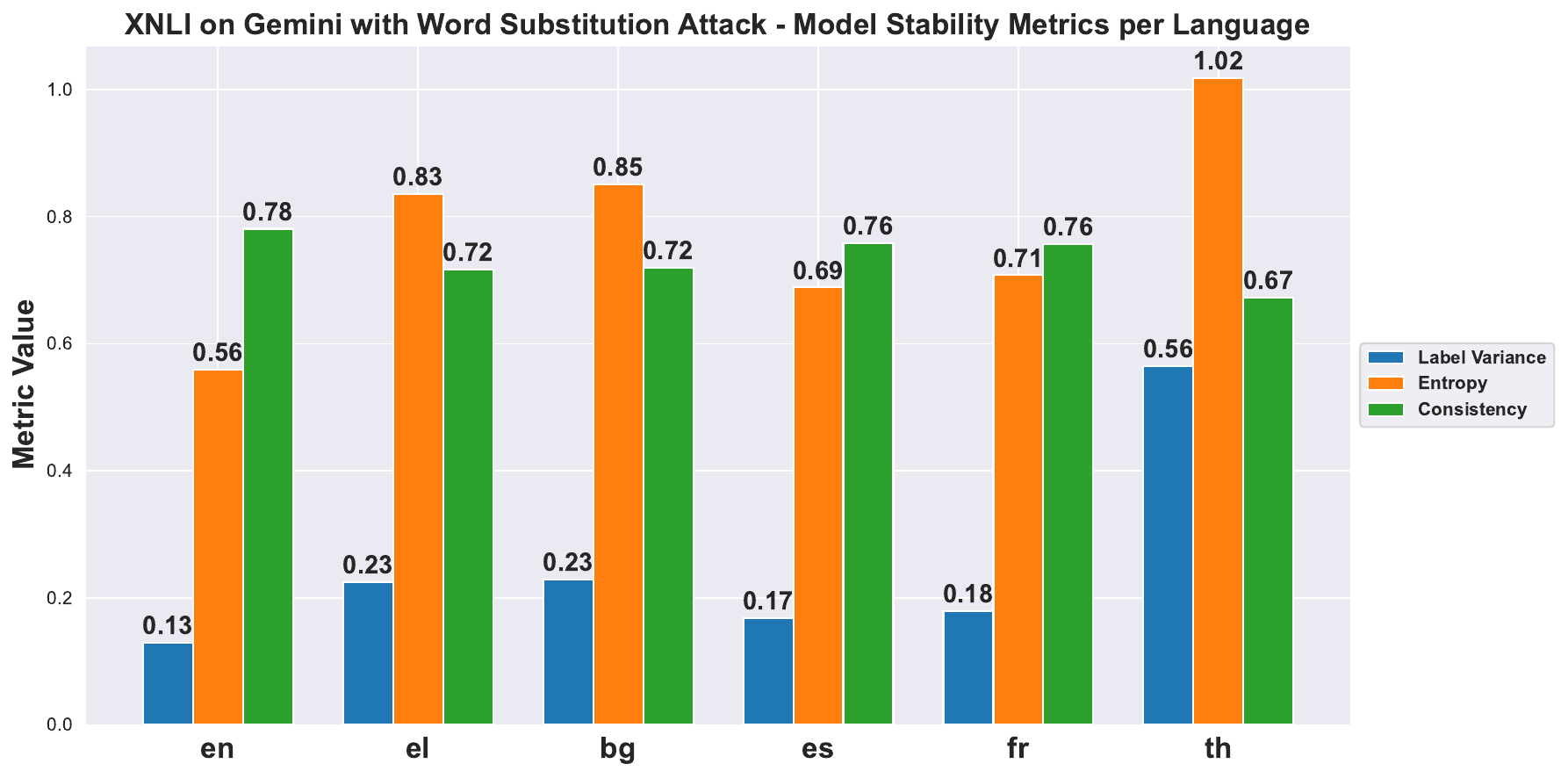}
    \caption{
    Stability metrics per language on the XNLI dataset with Gemini 1.5 Flash, under a BERT-based word substitution attack.
    Label variance reflects variability in predicted class labels across runs. 
    Entropy measures prediction uncertainty across labels. 
    Consistency indicates the proportion of predictions that agree with the majority vote. 
    These metrics characterize the robustness and prediction stability of the model under adversarial perturbation.
    }
    \label{fig:xnli_stability_attack}
\end{figure}

To evaluate the adversarial robustness of \textit{Gemini 1.5 Flash} on XNLI, we apply a BERT-based contextual word substitution attack, implemented using the \texttt{nlpaug} library, on the same data points previously evaluated under standard conditions. Specifically, we use the \texttt{ContextualWordEmbsAug} augmenter with language-specific Bert models, operating in substitution mode. Each input is perturbed with a probability of 15\% per token (\texttt{aug\_p = 0.15}), replacing words with semantically similar alternatives based on contextual embeddings.

Figure~\ref{fig:xnli_accuracy_attack} presents the mean accuracy across languages after the attack, along with the standard deviation, which captures variability in per-example correctness (1 = correct, 0 = incorrect) across runs. Although the magnitude of performance degradation varies across languages, the final accuracies after the attack are relatively close, with only a 4 percentage point difference between the highest- and lowest-performing languages. Notably, the standard deviation remains relatively low across all languages, suggesting that the degradation is systematic rather than erratic.

However, the above standard deviation reflects only variability in correctness, not in the model's actual label choices. To capture a more fine-grained view of prediction behavior, Figure~\ref{fig:xnli_stability_attack} reports standard stability metrics (label variance, entropy, and consistency) after the attack. These reveal that, although overall accuracy is similar across languages, the model’s response stability varies substantially. English demonstrates the highest stability across all three metrics, particularly in entropy and label variance. In contrast, Thai exhibits the highest entropy (1.02) and label variance (0.56), alongside the lowest consistency (0.67), indicating significantly increased prediction instability.

Further insight can be drawn from the confusion matrices in Figure~\ref{fig:confusion_matrices_xnli_attack} in Appendix~\ref{appendix:graphs}, which reveal that, after the attack, the model tends to overwhelmingly favor the \textit{neutral} label across most languages. The main exception is Thai (th), which shows a more balanced distribution across labels, even if most of the predictions are wrong. This defaulting to neutrality contributes to the relatively uniform accuracy across languages, despite varying stability and label distributions. This behavior reinforces prior observations that, under uncertainty, LLMs often default to “safer” or less committal outputs, in this case, neutral classifications, rather than risking stronger, potentially incorrect inferences. 

These results highlight a vulnerability of \textit{Gemini} to adversarial attacks across all evaluated languages. However, they also underscore a notable asymmetry in robustness: the model is significantly more stable in English than in other languages, pointing to its stronger internal representation and confidence when operating in its training-dominant language.
\newpage
\paragraph{Evaluation with LLaMA 3.2–3B}\mbox{}\\

\begin{figure}[H]
    \centering
    \includegraphics[width=0.9\linewidth]{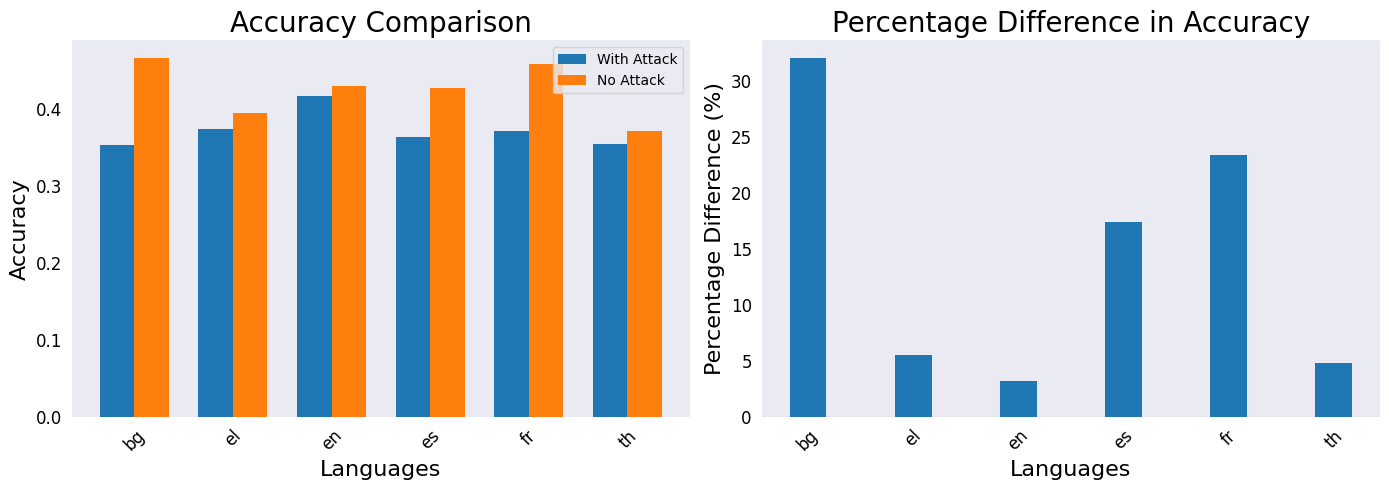}
    \caption{Accuracies per language on the XNLI dataset with LLaMA 3.2-3B, before and after applying the word substitution attack. The left figure shows the accuracies and the right one the percentage drop after the attack.}
    \label{fig:llama_accuracies_attack}
\end{figure}

To evaluate the robustness of LLaMA in each language we use the XNLI dataset and perform two experiments, one with the actual data points and one with the word substitution attack. In figure \ref{fig:llama_accuracies_attack} we observe the accuracies per language and how their performance is affected by the attack. From the percentage difference graph it is clear that the model is more consistent and reliable in English than all other used languages. A surprising result here is for Spanish and French, languages similar to English, the model seems to not be as robust, while in lower resource languages, such as Greek and Thai, it is.
\subsubsection{XQuAD}
\begin{figure}[H]
    \centering
    \includegraphics[width=0.6\linewidth]{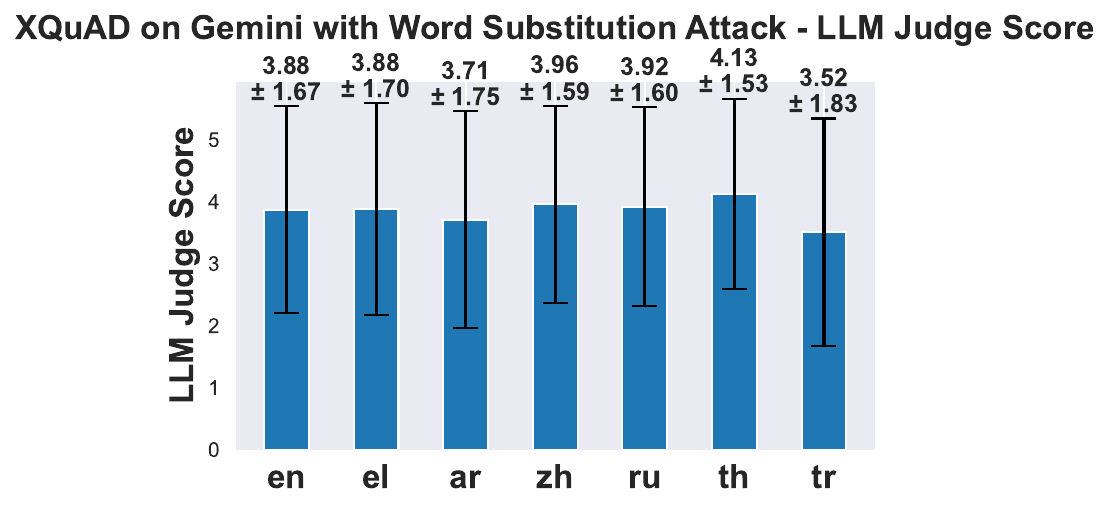}
    \caption{Mean LLM Score (on a 1–5 scale) per language on the XQuAD dataset with Gemini 1.5 Flash, after applying a BERT-based word substitution attack. Error bars show standard deviation across examples.}
    \label{fig:xquad_score_attack}
\end{figure}

To evaluate the adversarial robustness of \textit{Gemini 1.5 Flash} on XQuAD, we apply a BERT-based contextual word substitution attack, implemented using the \texttt{nlpaug} library, on the same data points previously evaluated under standard conditions. Specifically, we use the \texttt{ContextualWordEmbsAug} augmenter with language-specific Bert models, operating in substitution mode. Each input is perturbed with a probability of 30\% per token (\texttt{aug\_p = 0.30}), replacing words with semantically similar alternatives based on contextual embeddings.

Figure~\ref{fig:xquad_score_attack} reports the mean LLM Judge Scores across languages after the word substitution attack. Additional metrics (BLEU, METEOR, and cosine similarity) are provided in Figure~\ref{fig:xquad_metrics_attack} in Appendix~\ref{appendix:graphs}. However, due to the high variance observed in these automated metrics, we rely primarily on the LLM Judge Score as a more reliable and human-aligned evaluation metric.

Following the word substitution attack, the mean LLM Judge Scores across languages decreases, as shown in Figure~\ref{fig:xquad_score_attack}. While the post-attack scores remain relatively close, ranging from 3.52 to 4.13, the drop is evident across all languages. Thai (th) now achieves the highest average score (4.13), slightly ahead of Russian (ru) and Chinese (zh), whereas Turkish (tr) records the lowest score (3.52) and the highest variance ($\pm$1.83), indicating reduced reliability. English (en), which previously achieved near-perfect performance (4.94), drops to 3.88 with increased variance ($\pm$1.67).

Although these scores are still relatively strong given the adversarial nature of the attack, it is important to note that the task itself remains simple, requiring only the generation of basic answers to straightforward questions. Therefore, any significant drop in performance or stability is notable, as it highlights vulnerabilities in the model's robustness even in low-complexity, non-domain-specific settings.

The relatively uniform post-attack performance contrasts with the pre-attack results, where English clearly leads and variance is generally lower. This suggests that the word substitution attack affects all languages, but also disproportionately narrows the performance gap. These findings indicate that while Gemini handles simple QA tasks well under normal conditions, its resilience varies under adversarial input, and its strong performance in English becomes less pronounced in the presence of input perturbations.

\subsection{Cross-Linguistic Comparison}
We complete this research with an evaluation of language similarity and comparison of model performance based on this metric. To be more specific, we need to understand how different linguistic factors affect model performance, in order to get a clearer picture of the multilingual capabilities of LLMs.

To do this, we utilize lang2vec~\cite{littell2017uriel}, a resource that represents languages as vectors based on a variety of linguistic features such as syntax, phonology, and geography. These vectors allow us to quantitatively compare languages by capturing their typological properties. From lang2vec, we focus on two similarity measures: syntactic similarity and an averaged similarity, where we combine all available distance types (genetic, featural, geographic, syntactic, phonological, and phonetic) and then average. In practice, we rely on the precomputed pairwise distance matrices provided in lang2vec, which encode distances between languages on a normalized $[0,1]$ scale. We then derive similarity to English as $1 - \text{distance}$, following the convention used in the lang2vec paper.

We also use the World Atlas of Language Structures (WALS)~\cite{wals_online} as an alternative to derive syntactic similarities. Here, the scores are constructed directly from typological descriptors by comparing the presence or absence of selected features between each language and English. Specifically, we combine features 81A, 85A, 86A, 87A, 88A, 89A, 12A, and 50A, which capture aspects of word order, morphological marking, and clause structure. 

We use these similarity measures to assess how linguistic distance from English correlates with model performance across different languages, thereby gaining a clearer understanding of how typological differences impact multilingual model capabilities. We select English as the reference point because it is both the dominant training language for most large language models~\cite{liQuantifyingMultilingualPerformance2024} and the language in which nearly all datasets we evaluate have the most consistent coverage. Moreover, our hypothesis is that English dominates in terms of performance and that languages more similar to it will yield more accurate results. These make English the most natural baseline for comparison. However, we note that this choice is not the only possible baseline. Alternative approaches, such as measuring pairwise similarities among all languages, or using each language’s closest typological neighbor as a reference, could provide complementary insights into cross-lingual transferability. We leave such analyses for future work but highlight that our English-centered comparison aligns with how LLMs are predominantly developed today.

We first use syntactic similarity of languages to English, as provided by lang2vec, to assess their aggregated performance across all tasks on Gemini 1.5, which is the model evaluated on nearly all tasks (all except one) in our study. This allows us to observe the overall trend across the combined set of tasks. Next, to analyze the correlation between similarity to English and adversarial robustness, we focus on the XNLI dataset and use stability measures to examine patterns with respect to syntactic similarity in lang2vec. Finally, to incorporate alternative definitions of linguistic similarity, we analyze LLaMA’s performance on XNLI using both syntactic similarities derived from WALS features and the averaged similarities computed from lang2vec.

\paragraph{Evaluation with Gemini 1.5 Flash}\mbox{}\\

In order to assess each language’s correlation to performance and its similarity to English, 
we compute an aggregated score across all tasks and datasets for each language, using the same min-max normalization procedure 
described in Section~\ref{baseline_evaluation}. 
Here, however, we restrict the analysis to Gemini~1.5 Flash only, without averaging across models, since that is the model we used for most of the tasks. This normalization/aggregation  lets us directly compare languages even though the tasks and metrics differ, yielding an interpretable overall ranking of language robustness 
and performance.

While this approach provides a convenient way to aggregate heterogeneous metrics, 
it has certain drawbacks. 
First, relative differences are amplified: even when two languages perform almost equally well, 
the one that is slightly behind may receive a much lower normalized score. 
Second, when a dataset includes only a small number of languages (e.g., two), 
the min-max scaling collapses the comparison to a binary outcome, discarding finer distinctions. 
Thus, some languages may appear disproportionately stronger or weaker than expected. We therefore interpret the aggregated scores primarily as showing overall trends 
rather than as exact performance indicators.

\begin{figure}[H]
  \centering
  \includegraphics[width=\textwidth]{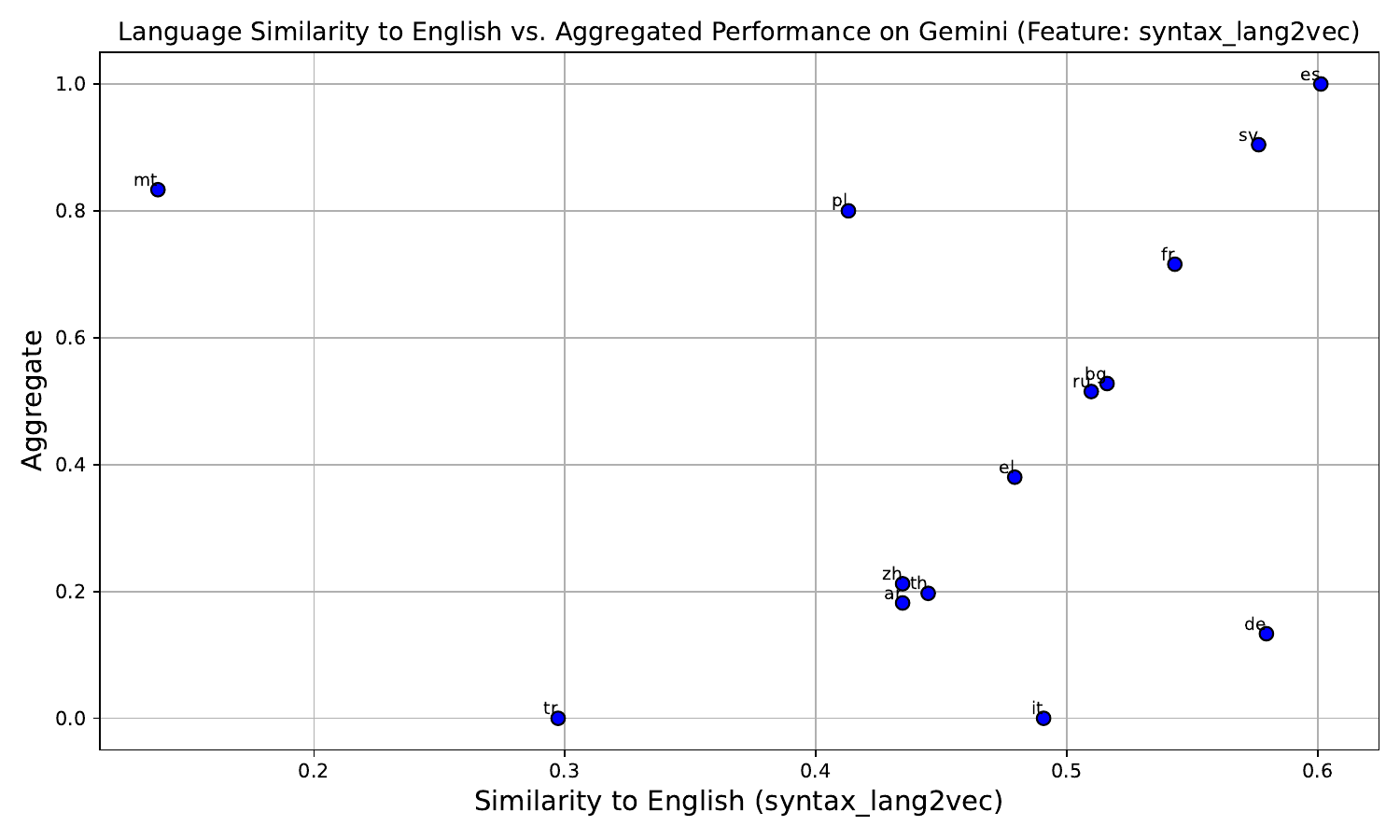}
  \caption{Aggregated per-language performance across all evaluated tasks for Gemini 1.5 Flash. 
  Scores are normalized per dataset using min–max scaling and then averaged across tasks.}
  \label{fig:aggregated_overview}
\end{figure}

Figure~\ref{fig:aggregated_overview} shows the aggregated scores per language against their syntactic similarity to English. A clear trend can be observed: there is an almost linear correlation between aggregated performance and similarity to English, with languages more similar to English generally achieving higher scores.  

Four languages appear as outliers to this trend: Maltese, Polish, Italian, and German. Three of these (Polish, Italian, and German) can be attributed to normalization effects and the limited representation of these languages in the datasets. Specifically, Italian and Polish only appear in the ToS dataset, where we use results from the highly assertive prompt. The scores in this dataset are very close (ranging from 0.61 to 0.66), but normalization exaggerates the difference: Italian (0.61) is heavily penalized while Polish (0.65) is disproportionately favored, even though the absolute gap is small. A similar situation occurs with German, which appears only in the ToS and LEXam datasets. In LEXam (MC and Open), German is compared only against English. Since German consistently performs slightly worse, it is normalized to 0, with no other information to balance the result. 

The only truly unexpected outlier is Maltese, which, despite being a relative uncommon language and having the lowest syntactic simialrity to English, achieves unexpectedly strong performance. This result aligns with earlier observations when comparing Maltese to English, where we already noted its performance as unexpectedly high. Our intuition is that Maltese may benefit from linguistic properties beyond syntactic similarity, which are not immediately captured by the similarity metric we employ. This suggests that while syntactic similarity is a strong general predictor of performance, other latent factors can also contribute to unexpected outcomes.

Nevertheless, the general trend shows that syntactic similarity to English correlates with overall model performance. Languages that share structural properties with English seem to benefit in performance, leading to consistently higher aggregated scores. This indicates that multilingual models still exhibit a strong bias towards languages that are closer to English in their syntactic profile, while more distant languages face a disadvantage, even when we include more specialized fields like law. In other words, the closer a language is to English syntactically, the more likely the model is to  perform better across tasks.

\begin{figure}[H]
    \centering
    \includegraphics[width=0.9\linewidth]{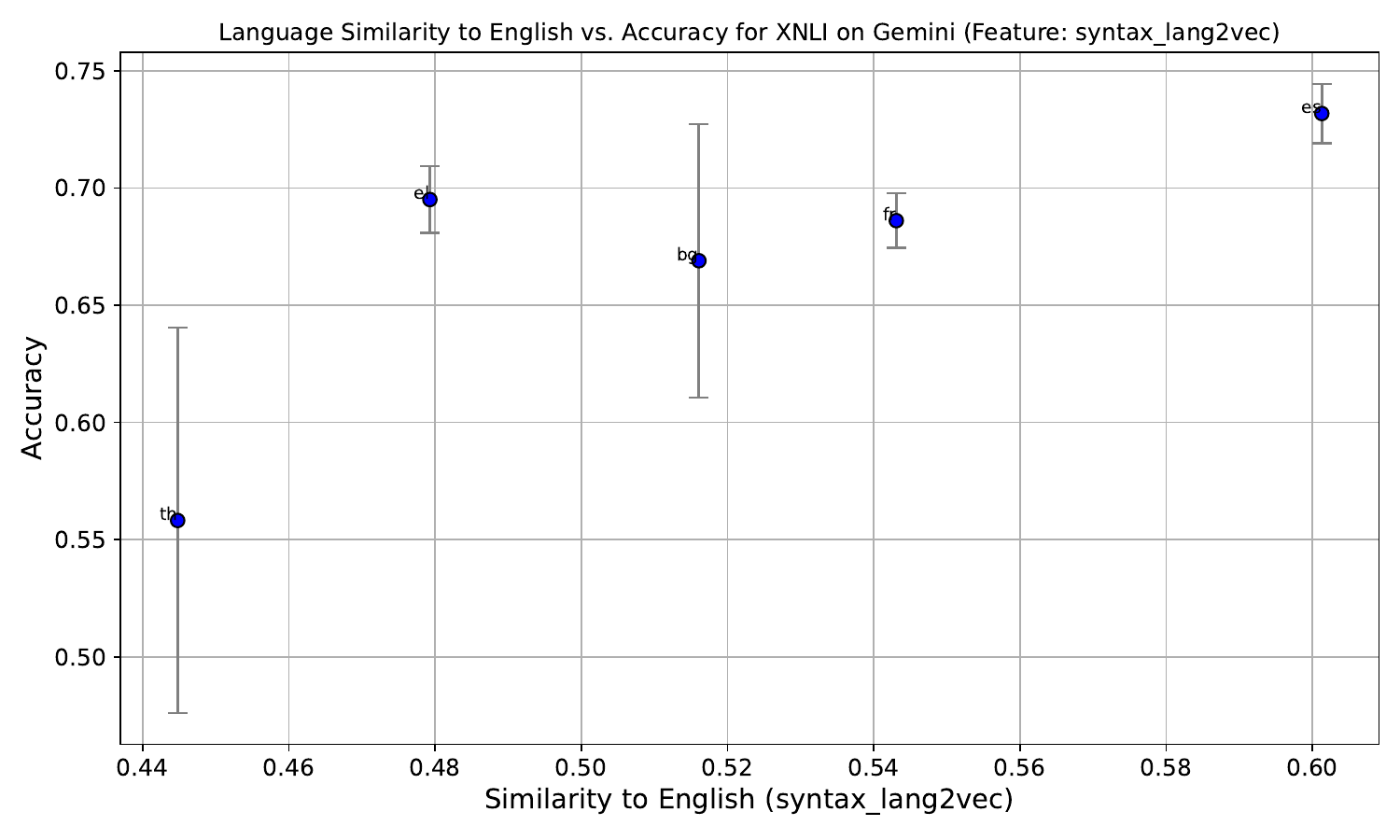}
    \caption{Accuracy per language on the XNLI dataset with Gemini 1.5 Flash, plotted against syntactic similarity to English. No attack applied. Error bars show standard deviation computed from per-example correctness (1 = correct, 0 = incorrect) across 25 runs, reflecting variability in predictions across runs.}
    \label{fig:accuracy_vs_similarity_gemini}
\end{figure}

To further assess the role of similarity to English, we evaluate on the XNLI dataset, which allows us to also incorporate the adversarial results. In addition to the adversarial setting, we also consider the standard (non-adversarial) XNLI results as a concrete example, rather than relying on the normalized aggregated scores. 

Figure~\ref{fig:accuracy_vs_similarity_gemini} shows the correlation between mean XNLI accuracy over 25 runs and syntactic similarity to English. Languages that are more syntactically similar to English (e.g., Spanish) tend to exhibit higher accuracy, while less similar ones (e.g., Thai) perform worse. Most languages appear to follow a linear relationship between accuracy and similarity, with the exception of Greek, which deviates from this trend. Although not definitive, this pattern suggests that syntactic similarity may modestly influence multilingual model performance.

\begin{figure}[H]
    \centering
    \includegraphics[width=0.9\linewidth]{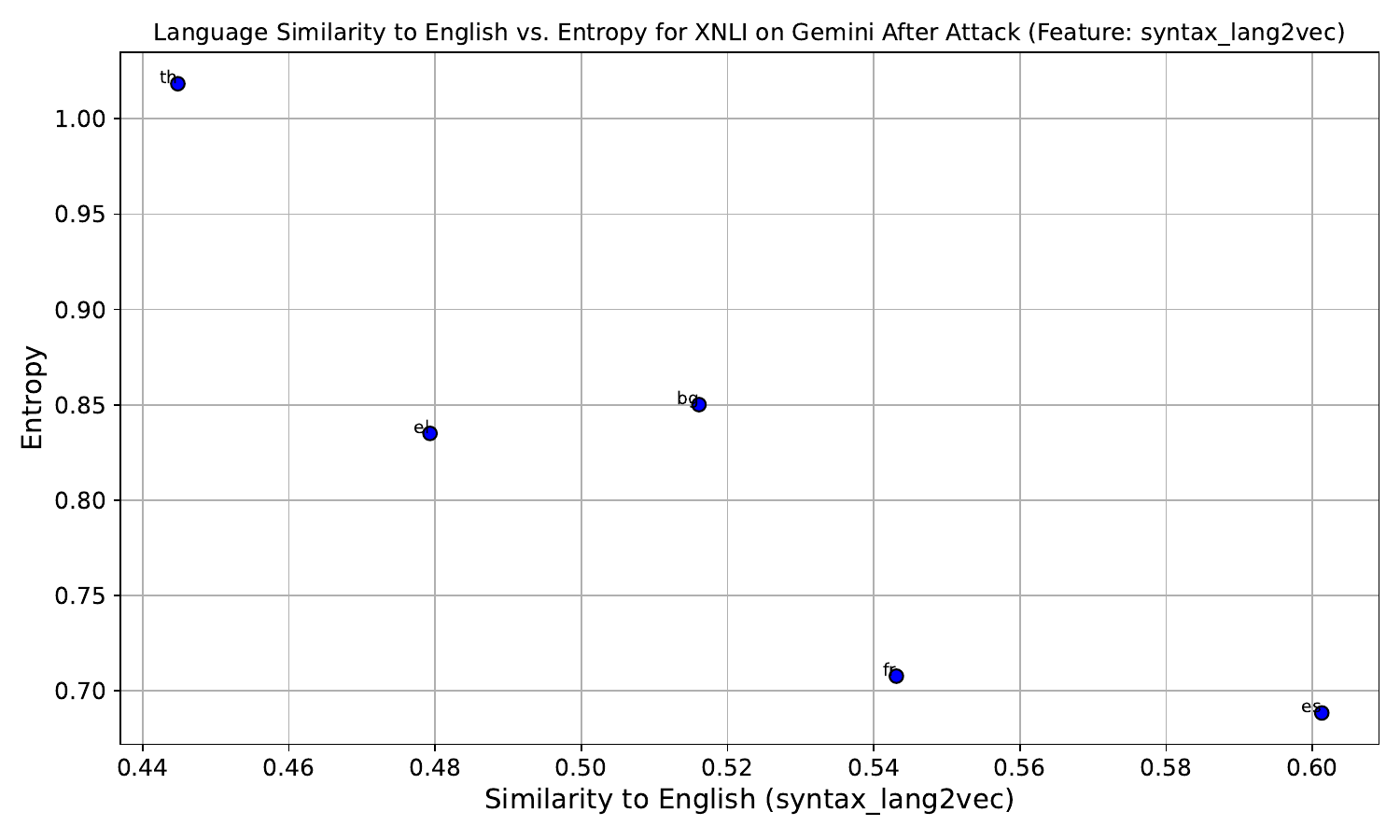}
    \caption{Shannon entropy per language on the XNLI dataset with Gemini 1.5 Flash, after applying a BERT-based word substitution attack. Higher values indicate greater uncertainty in model predictions.}
    \label{fig:entropy_vs_similarity_attack_gemini}
\end{figure}

\begin{figure}[H]
    \centering
    \includegraphics[width=0.9\linewidth]{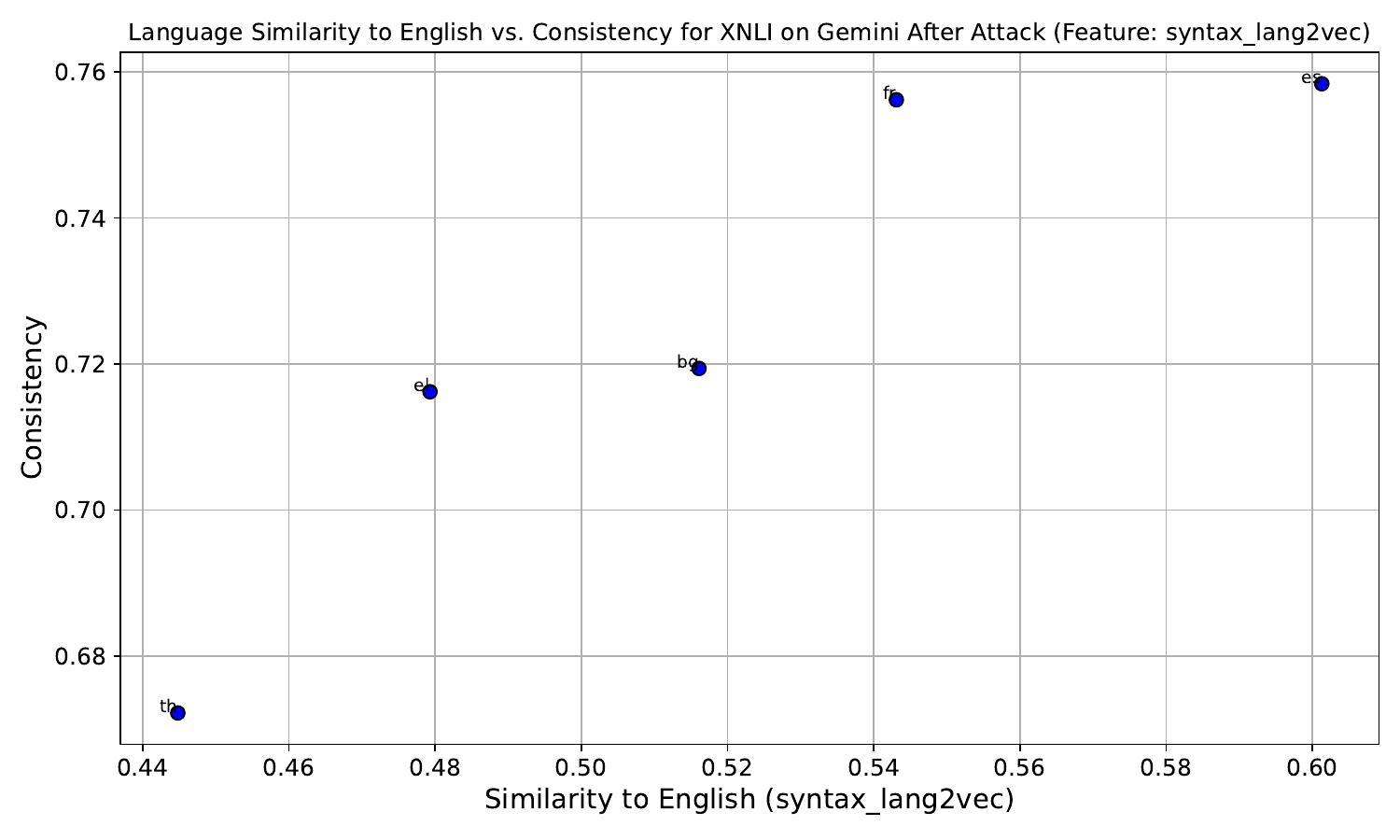}
    \caption{Prediction consistency per language on the XNLI dataset with Gemini 1.5 Flash, after applying a BERT-based word substitution attack. Consistency measures agreement between predictions.}
    \label{fig:consistency_vs_similarity_attack_gemini}
\end{figure}

For the post-attack results, we rely on stability-based metrics for evaluation, since accuracy values converge and become less informative. Figure~\ref{fig:entropy_vs_similarity_attack_gemini} presents the Shannon entropy of label predictions across 25 runs, while Figure~\ref{fig:consistency_vs_similarity_attack_gemini} shows consistency scores. Both metrics exhibit a clear linear trend: languages more similar to English tend to have lower entropy and higher consistency. Once again, Greek is a slight outlier since its entropy is marginally (0.015) lower than Bulgarian’s, despite Bulgarian being more similar to English. However, this difference is minimal and likely not meaningful.

Overall, these results indicate a consistent correlation between syntactic similarity to English and Gemini’s performance. While the evidence is not conclusive, the observed patterns suggest that languages closer to English tend to achieve higher accuracy and greater robustness, even under adversarial conditions where model stability becomes more critical. At the same time, the presence of occasional outliers highlights that syntactic similarity is not the sole factor driving multilingual performance, and other influences may also play an important role.

\paragraph{Evaluation with LLaMA 3.2–3B}\mbox{}\\
\begin{figure}[H]
    \centering
    \begin{subfigure}[b]{0.48\linewidth}
        \centering
        \includegraphics[width=\linewidth]{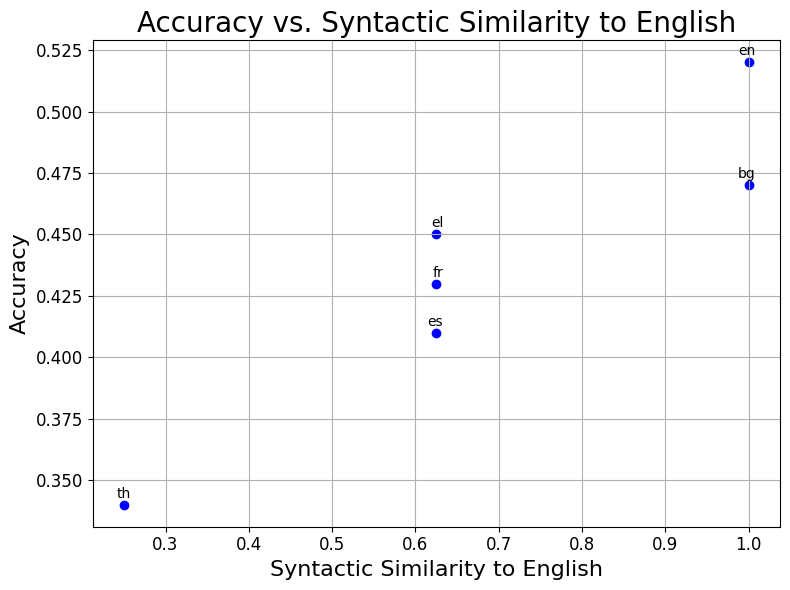}
        \caption{Syntactic similarity}
        \label{fig:llama_syntactic_similarity}
    \end{subfigure}
    \hfill
    \begin{subfigure}[b]{0.48\linewidth}
        \centering
        \includegraphics[width=\linewidth]{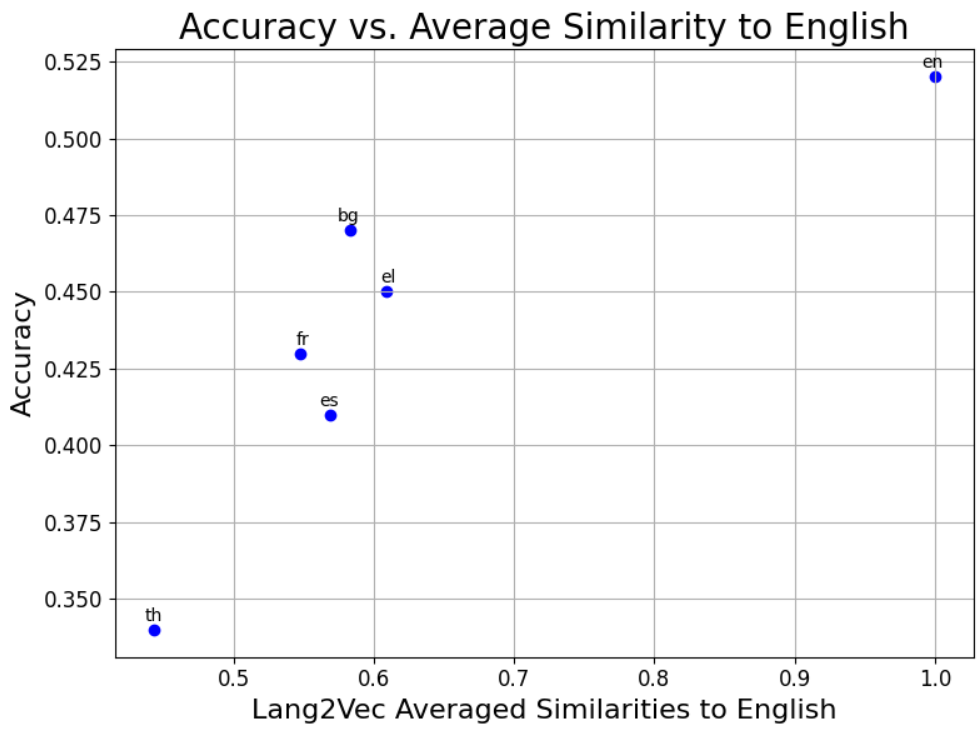}
        \caption{Averaged lang2lec similarity}
        \label{fig:llama_lang2vec_similarity}
    \end{subfigure}
    
    \caption{Comparison of LLaMA model similarities: (a) Syntactic and (b) Averaged lang2vec.}
    \label{fig:llama_similarity_combined}
\end{figure}

To evaluate the cross-linguistic performance of LLaMA, we analyze the relationship between its accuracy on the XNLI dataset and the linguistic similarities of each language to English. We use syntactic similarity using a combination of relevant features from WALS as well as pre-computational distances from lang2vec. 

To investigate the impact of linguistic similarity, we incorporate syntactic similarity scores derived directly from WALS features. Unlike the syntactic similarity used in the Gemini evaluation (which was based on pre-computed lang2vec vectors), here we construct the measure from WALS typological descriptors, focusing on properties such as word order, morphological marking, and clause structure. This provides a feature-level perspective on cross-linguistic relatedness that allows us to contrast results across different definitions of syntactic similarity.

Figure~\ref{fig:llama_syntactic_similarity} shows a linear relationship: the least similar language (Thai) yields the lowest accuracy, while the most similar (Bulgarian) achieves the highest. The remaining three languages, which share the same syntactic similarity score to English, fall in the middle range. These results offer a consistent correlation between model performance and linguistic similarity, reinforcing the findings from Gemini's cross-linguistic evaluation.

Furthermore, figure~\ref{fig:llama_lang2vec_similarity} presents combined similarity scores, derived from lang2vec, plotted against model accuracy. A positive correlation is observed, suggesting that LLaMA performs better in languages that are in general closer to English. However, the trend is not definitive, as some languages more similar to English exhibit lower accuracy than less similar ones. This discrepancy may be explained by other linguistic or dataset-specific factors, or by the averaging procedure itself: since all distance types in lang2vec are given equal weight, the influence of less relevant similarities may obscure the role of those that matter most for model performance.

While the sample size of languages remains limited, this analysis highlights the influence of linguistic alignment on LLaMA's multilingual performance and suggests an inherent bias toward English. This finding is consistent with prior work by Zhang et al.~\cite{zhang2023donttrustchatgptquestion}, who argue that GPT-3.5 exhibits subordinate multilingualism, relying heavily on English-centric reasoning even when responding in other languages.

\section{Discussion}
The evaluation across multilingual datasets reveals that current LLMs, while advancing rapidly, still exhibit clear limitations, especially under adversarial conditions or complex legal reasoning tasks. English consistently shows higher stability, suggesting training data bias or architectural alignment with English syntax and semantics. However, in some benchmarks, even if limited, other languages outperform English in raw accuracy, indicating that performance does not always correlate with resource availability. Nevertheless, in simple, non-adversarial settings, English generally dominates in both performance and stability. This outcome is not unexpected, as prior work outside the legal domain also supports such trends~\cite{hu2020xtreme, zhang2023donttrustchatgptquestion}.

Adversarial robustness remains a major challenge. While syntactic (character-level) attacks in our experiments do not substantially affect performance, semantic (word-substitution) attacks reduce it, often dramatically, especially in open-ended tasks like legal reasoning (LEXam), consistent with prior work on general datasets showing that semantic perturbations are typically more damaging than character-level ones~\cite{wang2024decodingtrustcomprehensiveassessmenttrustworthiness}. We also observe that when the attack strength is relatively high, performance across languages tends to converge, effectively overcoming the relative advantage of English. Unexpectedly, in some cases, the percentage drop in English performance after an attack is larger than that of other languages, though this is not always the case. Nevertheless, English still yields consistently more stable results in terms of reproducibility across runs (i.e., giving the same answer for the same question), thereby showcasing higher trustworthiness in the outputs for English tasks. 

Moreover, syntactic similarity to English for a given language seems to have a level of correlation with model performance in that language. More specifically, languages that are more syntactically similar to English tend to show a general improvement in performance. In adversarial settings, these languages also exhibit lower entropy across different runs and higher consistency, indicating that models produce more stable and dependable outputs when the syntax of the input language closely resembles that of English. This further confirms the anglophone bias of even multilingual LLMs and aligns with prior non-legal research showing that similarity to English is often predictive of stronger cross-lingual transfer and more reliable model behavior. Such recent work~\cite{xuCrossLingualPitfallsAutomatic2025} demonstrates that, in general-purpose datasets, linguistically similar languages tend to share performance patterns. Similarly, studies on multilingual BERT (mBERT)~\cite{kCrossLingualAbilityMultilingual2020} highlight the importance of structural similarity for cross-lingual ability. Complementing this, recent research~\cite{eronenZeroshotCrosslingualTransfer2023} shows that linguistic similarity, when measured as a combination of typological features, correlates with transfer performance between languages.

At the same time, our analysis also reveals notable outliers to this near-linear trend, suggesting that syntactic similarity is not the sole driver of performance. A few languages may deviate from the expected correlation, where despite low syntactic similarity to English, they achieve comparatively strong results. This indicates that other factors beyond syntax, perhaps different linguistic characteristics or general availability of the language, can play an important role.  Moreover, as a recent study highlights~\cite{samardzicMeasureTransparentComparison2024}, benchmark datasets themselves may under- or miss-represent low-resource languages, introducing biases. This implies that part of the observed correlation could be amplified by dataset composition rather than reflecting purely model-internal properties. Overall, while syntactic similarity to English is a strong predictor of performance, these outliers and dataset limitations highlight the need to consider additional linguistic and resource-based influences when assessing cross-lingual capabilities of LLMs.

Importantly, Gemini 2.5 Flash demonstrates clear improvements over its predecessor, Gemini 1.5 Flash, in both robustness and overall performance, with the most pronounced gains observed in English. At the same time, our evaluation indicates that these advances come with certain trade-offs: while performance levels rise, the newer model can also display occasional inconsistencies in output formatting and language use, something not seen in its predecessor's answers. Such behaviors, though not directly undermining accuracy, highlight that progress in model capability does not always translate into more predictable or controllable behavior, a point particularly relevant when considering deployment in high-stakes legal contexts.

Finally, the study demonstrates that model behavior is highly sensitive to prompt formulation. This is particularly evident in fairness classification tasks (ToS dataset), where more assertive prompts significantly improve performance and reduce misclassification penalties, most notably in English. This reveals a conservative behavior in general-purpose LLMs, which may opt for neutral, legally safer or more positive answers when faced with ambiguity. This is in line with prior research outside the legal domain, which shows that LLMs often express uncertainty when confronted with ambiguous scenarios~\cite{scherrerEvaluatingMoralBeliefs2023}, and that models trained predominantly on public web content may exhibit a tendency to avoid accusatory or critical language~\cite{salechaLargeLanguageModels2024}, thereby defaulting to more positive responses.

\section{Conclusion}
This study provides a comprehensive assessment of the performance and robustness of modern Large Language Models (LLMs), specifically Meta’s LLaMA and Google’s Gemini, in multilingual and legal contexts. The results reveal that while English often offers more stable outcomes, it does not always guarantee higher accuracy. The models exhibit varying levels of effectiveness across tasks and languages, with performance generally correlating with syntactic similarity to English.

Despite recent advancements, significant challenges remain. Legal tasks, especially those requiring nuanced reasoning or fairness assessment, expose model weaknesses, particularly under adversarial conditions. The Gemini 2.5 model demonstrates notable improvements over its predecessor, yet struggling with prompt adherence and occasional language drift, raising concerns about reliability in high-stakes applications. Furthermore, we note that Gemini consistently outperforms LLaMA in our evaluations, suggesting that commercial models may be ahead in multilingual performance and robustness, though both still face serious limitations in high-stakes legal applications

Moreover, prompt design is found to significantly influence model behavior, especially in tasks like fairness classification, where assertiveness in instruction improved accuracy. This underlines the importance of thoughtful prompt engineering in mitigating model bias and ambiguity.

Finally, this study also highlights the limited adversarial robustness of current LLMs. Semantic perturbations, in particular, degrade performance markedly, underscoring the vulnerability of these systems in realistic, noisy input scenarios.

In summary, while LLMs are becoming increasingly capable, deploying them in multilingual, high-stakes domains like law requires caution. There is a clear need for continued research into improving model reliability, robustness, and fairness, especially for languages and legal systems underrepresented in current training data. To support this effort, we introduce an open-source, scalable evaluation pipeline that enables easy integration of new models and datasets, supports adversarial robustness testing, and includes an LLM-as-a-Judge module for semantically informed assessment.

\printbibliography

\appendix
\section{Languages Abbreviations}
\label{appendix:abbreviations}
\begin{table}[H]
\centering
\small
\begin{tabular}{|c|l|}
\hline
\textbf{Abbreviation} & \textbf{Language} \\
\hline
ar & Arabic \\
bg & Bulgarian \\
de & German \\
el & Greek \\
en & English \\
es & Spanish \\
fr & French \\
it & Italian \\
mt & Maltese \\
pl & Polish \\
ru & Russian \\
sv & Swedish \\
th & Thai \\
tr & Turkish \\
zh & Chinese \\
\hline
\end{tabular}
\caption{Mapping of ISO 639-1 language codes to full language names.}
\label{tab:lang_abbr}
\end{table}

\section{Additional Graphs}
\label{appendix:graphs}
\subsection{Online Terms of Service}

\begin{figure}[H]
    \centering
    \includegraphics[width=0.75\linewidth]{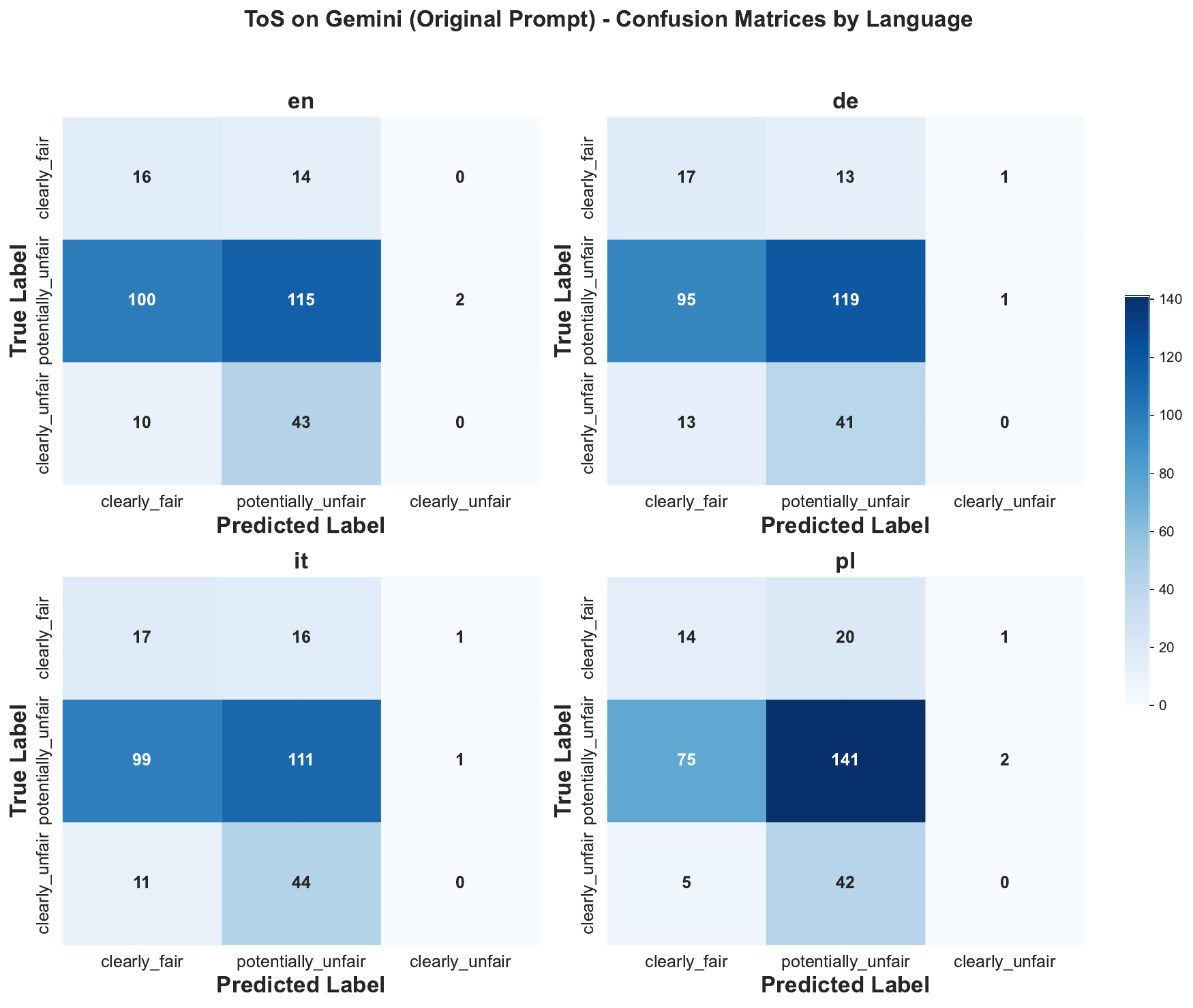}
    \caption{Confusion matrices (per language) on the Online Terms of Service dataset, with Gemini 1.5 Flash, using the basic prompt. The matrices reflect counts of predicted versus true labels for the three fairness classes: clearly fair, potentially unfair, and clearly unfair.}
    \label{fig:confusion_matrices_original}
\end{figure}

\begin{figure}[H]
    \centering
    \includegraphics[width=0.9\linewidth]{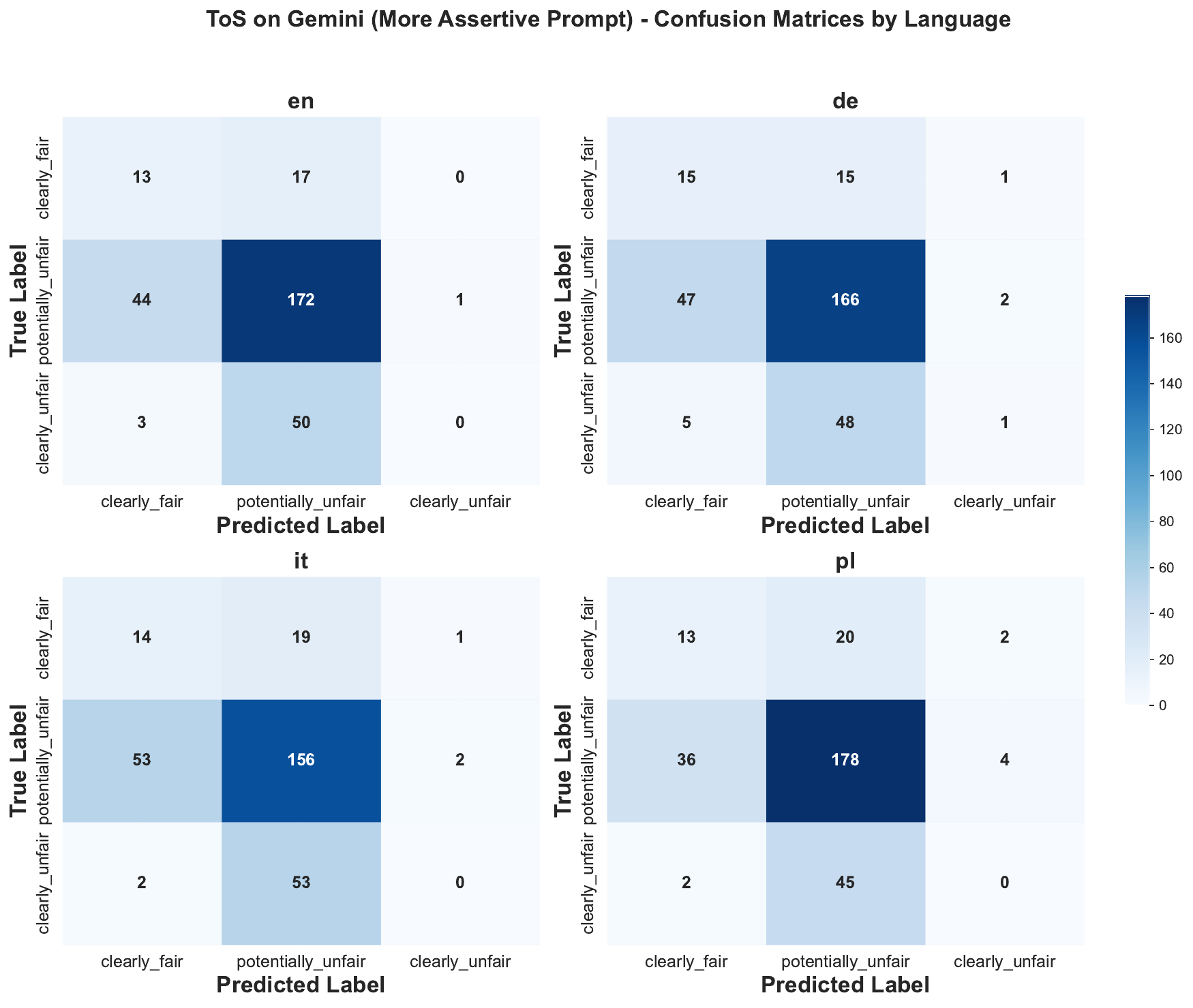}
    \caption{Confusion matrices (per language) on the Online Terms of Service dataset, with Gemini 1.5 Flash, using the more assertive prompt. The matrices reflect counts of predicted versus true labels for the three fairness classes: clearly fair, potentially unfair, and clearly unfair.}
    \label{fig:confusion_matrices_assertive}
\end{figure}

\begin{figure}[H]
    \centering
    \includegraphics[width=0.9\linewidth]{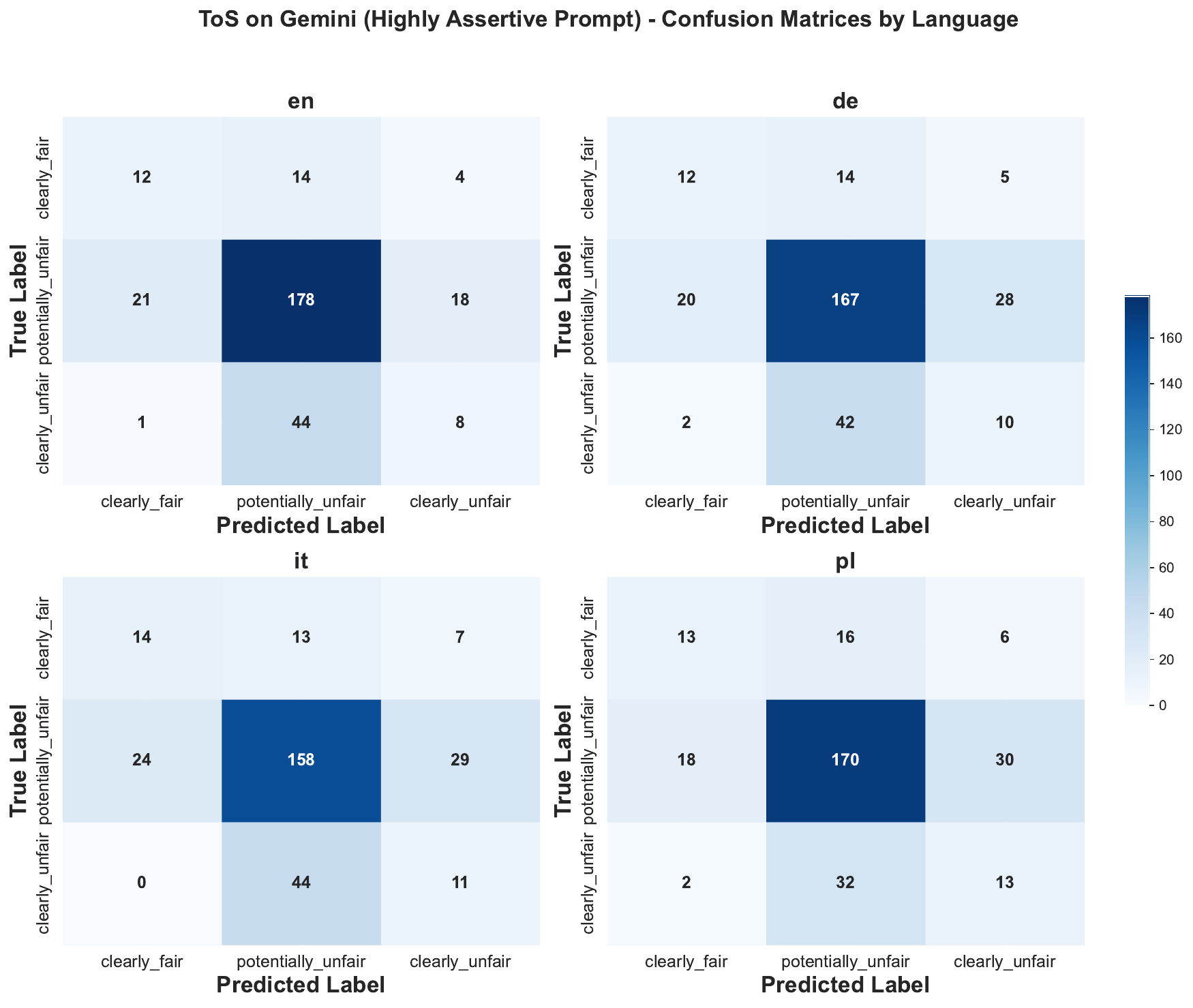}
    \caption{Confusion matrices (per language) on the Online Terms of Service dataset, with Gemini 1.5 Flash, using the highly assertive prompt. The matrices reflect counts of predicted versus true labels for the three fairness classes: clearly fair, potentially unfair, and clearly unfair.}
    \label{fig:confusion_matrices_highly_assertive}
\end{figure}

\subsection{XNLI}

\begin{figure}[H]
    \centering
    \includegraphics[width=0.9\linewidth]{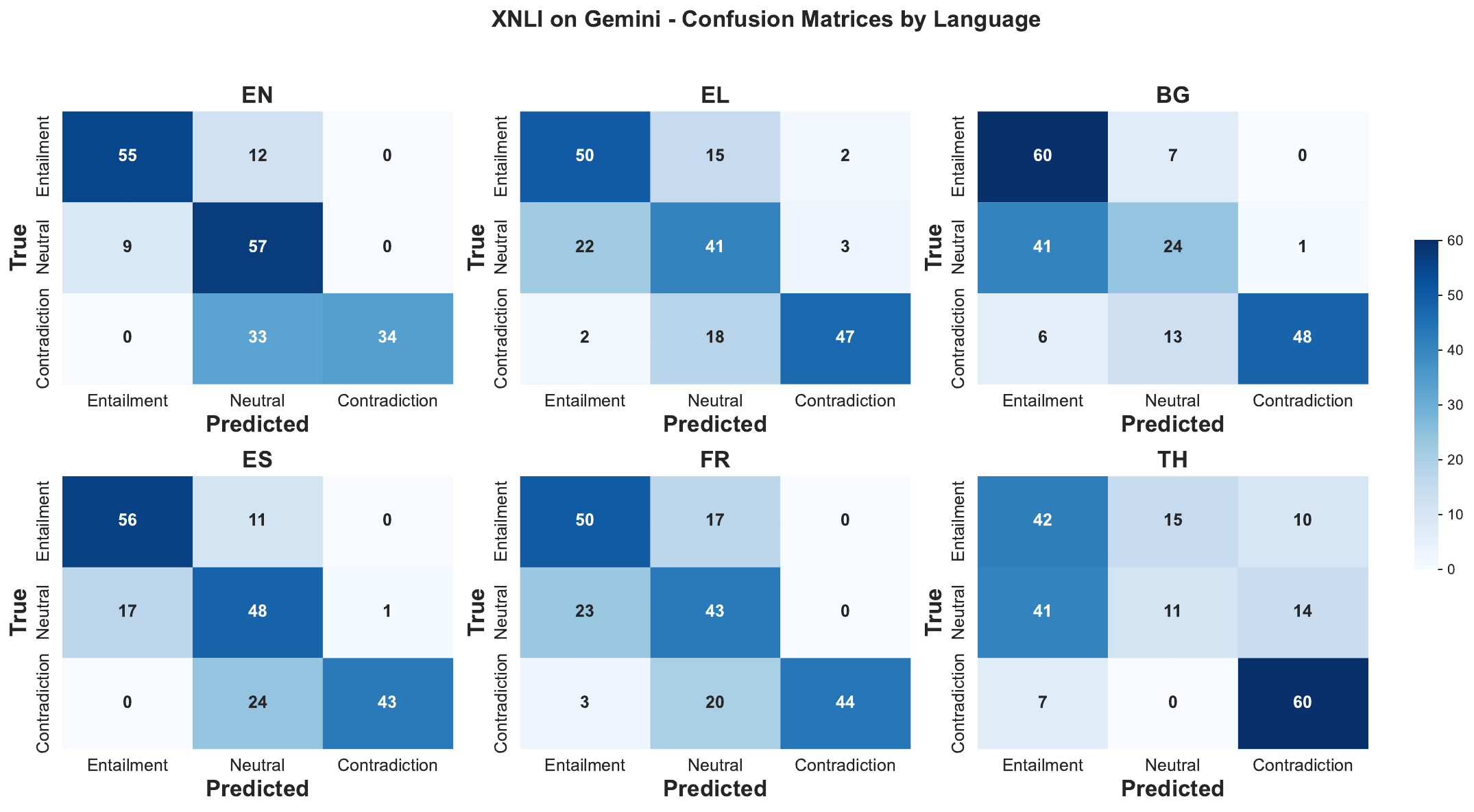}
    \caption{
    Confusion matrices per language on the XNLI dataset with Gemini 1.5 Flash. Each matrix shows counts of predicted versus true labels for the three inference classes: \textit{entailment}, \textit{neutral}, and \textit{contradiction}. These matrices highlight the distribution of prediction errors and classification tendencies across languages.
    }
    \label{fig:confusion_matrices_xnli}
\end{figure}

\begin{figure}[H]
    \centering
    \includegraphics[width=0.9\linewidth]{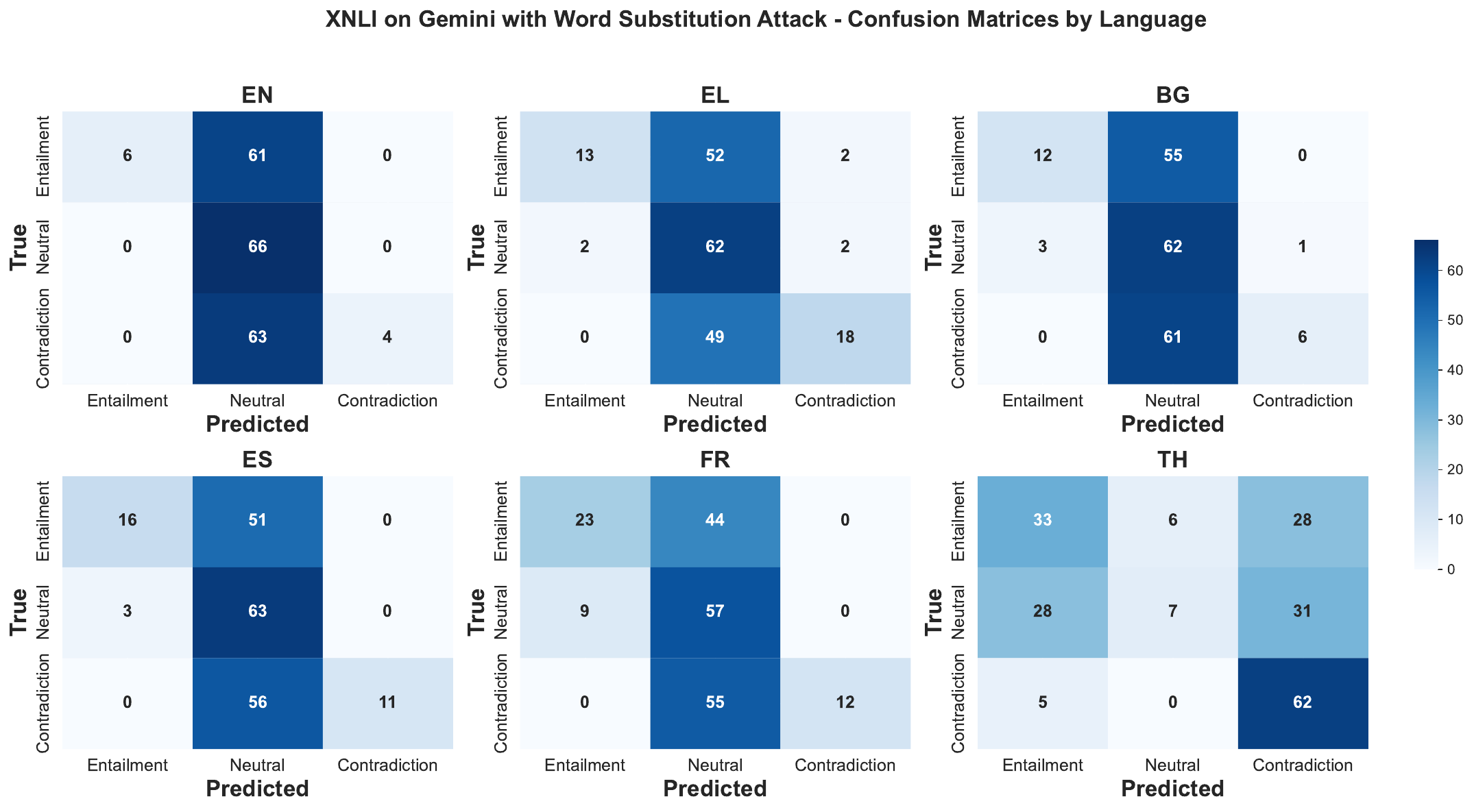}
    \caption{
    Confusion matrices per language on the XNLI dataset with Gemini 1.5 Flash after the BERT-based word substitution attack. 
    Each matrix shows counts of predicted versus true labels for the three inference classes: \textit{entailment}, \textit{neutral}, and \textit{contradiction}. The labels were assigned using majority vote across 25 runs.
    These matrices illustrate the effect of adversarial perturbations on prediction distributions and classification behavior across languages.
    }
    \label{fig:confusion_matrices_xnli_attack}
\end{figure}

\subsection{XQuAD}

\begin{figure}[H]
    \centering
    \includegraphics[width=0.9\linewidth]{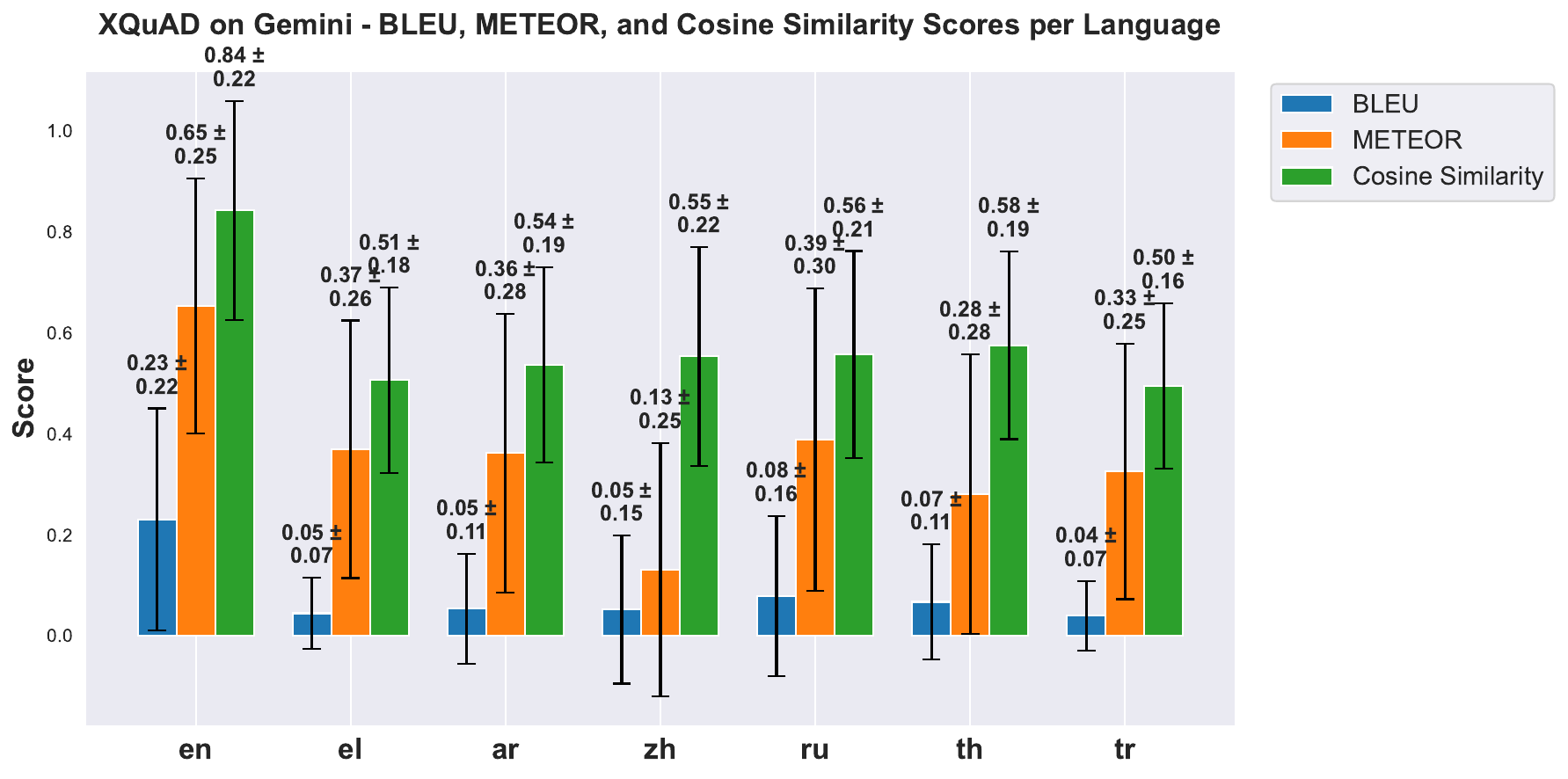}
    \caption{
    BLEU, METEOR, and Cosine Similarity scores per language on the XQuAD dataset with Gemini 1.5 Flash.
    Error bars indicate standard deviation across examples and reflect natural variability in generation quality.
    }
    \label{fig:xquad_metrics}
\end{figure}

\begin{figure}[H]
    \centering
    \includegraphics[width=0.9\linewidth]{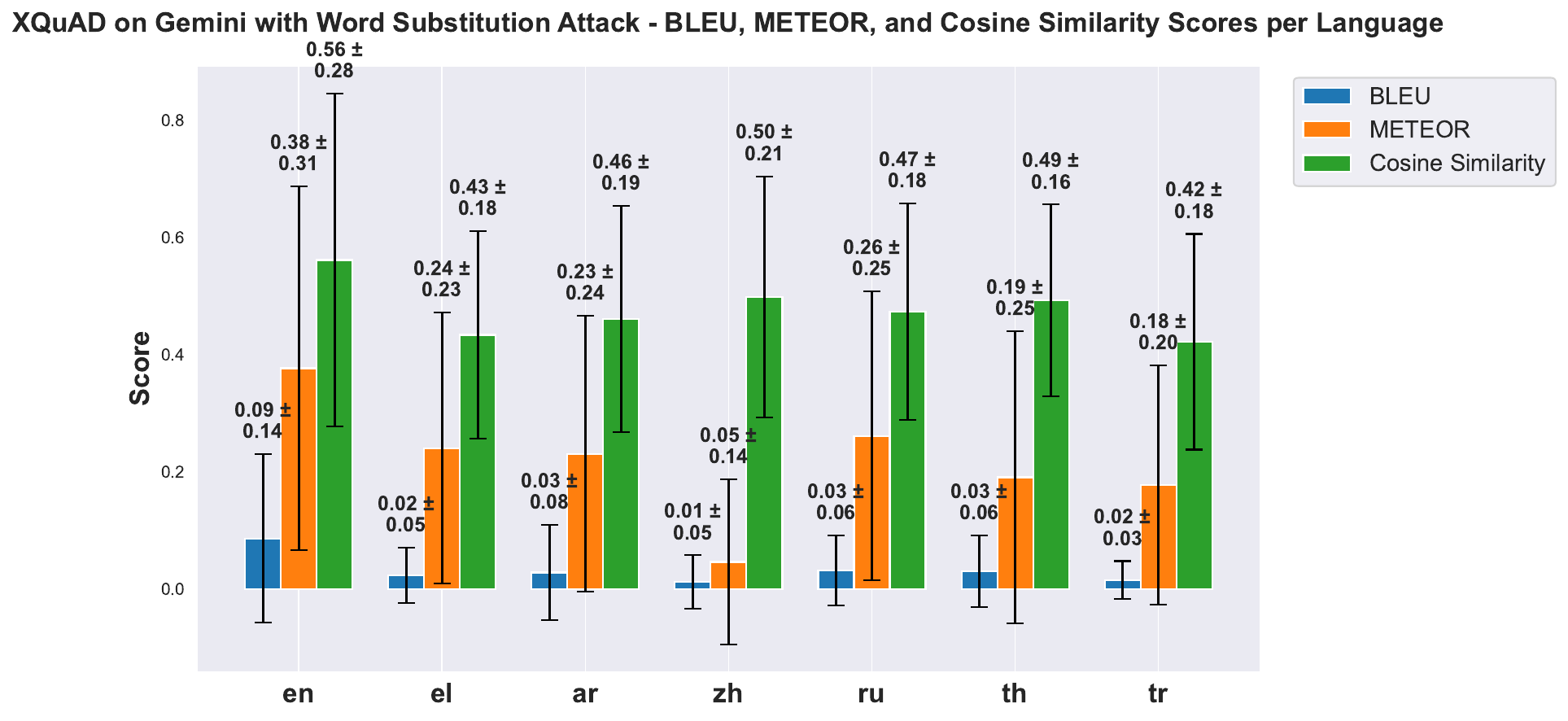}
    \caption{
    BLEU, METEOR, and Cosine Similarity scores per language on the XQuAD dataset with Gemini 1.5 Flash, after applying a BERT-based word substitution attack. 
    Error bars indicate standard deviation across examples and reflect variability in output quality under perturbation.
    }
    \label{fig:xquad_metrics_attack}
\end{figure}

\end{document}